\documentclass[11pt]{article}
\usepackage{hyperref}
\usepackage{amsmath, amsfonts, amssymb, mathrsfs}
\usepackage{dcolumn}
\usepackage{caption}
\usepackage{subcaption}
\usepackage{filemod}
\usepackage{natbib}
\usepackage{floatrow}
\usepackage{setspace}
\usepackage{verbatim}
\usepackage{graphicx}
\usepackage{url}
\usepackage{tikz}
\usepackage{bbm}
\usepackage{algorithm}
\usepackage{algpseudocode}
\usetikzlibrary{shapes.geometric, arrows}

\title{\vspace{-0.3cm} Gaussian Process Assisted Meta-learning for\\Image Classification and Object Detection Models}
\author{Anna R. Flowers\thanks{Corresponding author: \texttt{arflowers@vt.edu}} \thanks{Department of Statistics, Virginia Tech}  
	\and Christopher T. Franck\footnotemark[2]
	\and Robert B. Gramacy\footnotemark[2]
	\and Justin A. Krometis\thanks{Department of Mathematics, Virginia Tech} \thanks{Virginia Tech National Security Institute}}
\date{\today}

\begin{document}

\vspace{-0.5cm}
\maketitle
\singlespacing

\vspace{-1.5cm}

\begin{abstract} 
Collecting operationally realistic data to inform machine learning models can
be costly. Before collecting new data, it is helpful to understand where a
model is deficient. For example, object detectors trained on images of rare
objects may not be good at identification in poorly represented conditions. We
offer a way of informing subsequent data acquisition to maximize model
performance by leveraging the toolkit of computer experiments and metadata
describing the circumstances under which the training data was collected
(e.g., season, time of day, location). We do this by evaluating 
 the learner as the training data is varied according to its metadata. A
Gaussian process (GP) surrogate fit to that response surface can inform new
data acquisitions. This meta-learning approach offers improvements to learner
performance as compared to data with randomly selected metadata,
 which we illustrate on both classic learning examples, and on a
motivating application involving the collection of aerial images in search of
airplanes.

 \end{abstract}

\noindent \textbf{Keywords:}  computer experiment, metadata, data acquisition

\singlespacing

\section{Introduction}\label{sec:intro}

We consider scenarios where labeled data $D_N$ are used to train a machine
learning (ML) model (e.g., neural network
\citep[NN;][]{rosenblatt1958perceptron, rumelhart1986learning}, random forest
\citep[RF;][]{breiman2001random}, or support vector machine
\citep{cortes1995support}), but collecting and labeling data is the most
expensive part of the process. Examples of such scenarios include medical
imaging \citep{willemink2020preparing, hoi2006batch}, natural language
processing \citep{olsson2009literature}, speech recognition
\citep{abdelwahab2019active}, and streaming data \citep{vzliobaite2013active,
zhu2007active}. In these scenarios, it is important to ensure that the data
collected provides the ML model with as much information as possible.
Consider, for example, collecting images used to train an image classifier.
After training, the model will correctly classify some images and incorrectly
classify others. Our goal is to identify the {\em types} of images the model
categorizes accurately, or otherwise, in hopes of curating the composition of
future image collection.

Selecting data with the goal of improving a model is called active learning
(AL). The core idea behind AL is to choose new data points in an intelligent
way so that model performance is as good as possible under stringent data
budgets. There is an extensive literature of AL within the field of ML, which
we refer to as ML-AL; for a full review of ML-AL see \citet{settles2009active}
or \citet{ren2021survey}.  ML-AL typically assumes that unlabeled data is
cheap, and that only labeling is expensive. The process may involve requesting
labels for a chosen combination of inputs \citep[membership query
synthesis;][]{angluin1988queries} or for specific data points selected from a
pool of unlabeled ones \citep[stream- and pool-based
sampling;][]{lewis1995sequential, dagan1995committee}, whereby they are
subsequently labeled, say via uncertainty sampling
\citep{lewis1994sequential}, expected model change \citep{settles2007multiple,
roy2001toward}, diversity sampling \citep{brinker2003incorporating},
query-by-committee \citep{seung1992query}, or hybrids thereof
\citep{yang2015multi, ash2019deep,yin2017deep}. Both options have drawbacks.
Since the input space is complex for ML models, membership query synthesis may
request a combination of inputs that is unrecognizable to the annotator,
meaning it cannot be labeled \citep{baum1992query}.  Stream- and pool-based
sampling methods are far more common, but rely on the assumption that
collecting unlabeled data is cheap and only the labeling is expensive. When
collecting unlabeled data is also expensive, another AL method is needed.

Our idea is to select new data points according to their metadata. In this
paper we define metadata as any feature of the data that is not available to
the model during training. Since image classifiers, for example, are trained
using only the pixels of images, any extra information about those images can
be treated as metadata. We choose to select new data points in this way
because we believe that data points with different metadata may yield
different model performance. By choosing new data points according to their
metadata, we learn the types of data points that are most useful to the model.

\begin{figure}[ht!]
\centering 
\includegraphics[scale = 0.36]{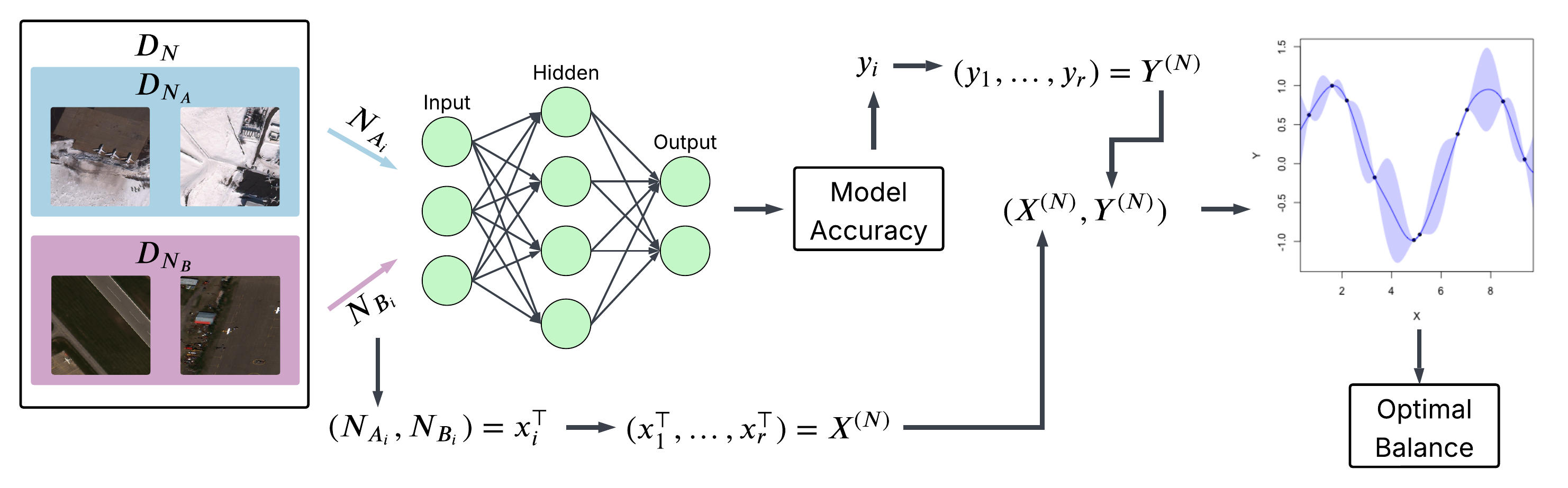}
\caption{Process for identifying the optimal balance of metadata.}
\label{fig:process}
\end{figure}

As an example, consider a motivating dataset known as RarePlanes
\citep{shermeyer2021rareplanes}, a collection of 8,525 aerial images of 25,205 planes.
RarePlanes contains many examples of metadata (e.g., weather, time of day,
location).  Exploring the relationship between these quantities and model
performance in the RarePlanes dataset is not a new idea
\citep{lanus2024coverage,torres2020odin}; however, we are the first
to frame this problem in an AL context.  Here we focus on a metadata record
indicating weather when the photo was taken: snowy, or not snowy.  To see why
data points with one type of weather metadata may be more valuable than another,
see Figure \ref{fig:process}.
For now, consider only the box labeled $D_N$ on the far left, representing the
entire dataset. Inside this box are two smaller boxes, colored blue and pink,
containing two images each. Blue and pink boxes represent two metadata
categories; blue with snow, and pink without. Since many of the planes are
white, they are harder to distinguish from the background in images with snow 
(verified in Appendix \ref{sup:rareplanesmeta}).
It is therefore harder for an ML object detector to accurately detect planes
in images with snow than otherwise. However, what this means about ideal data
for learning is not immediate. Harder to distinguish images may provide more
information per data point, but they risk decreasing predictive accuracy for
images without snow, which may be more common in applications where 
the model is deployed.


Somewhat more generically, let $A$ denote one metadata category (e.g., snow),
and $B$ denote its complement (not snow).  Throughout this paper we consider
only two metadata categories and discuss potential for more categories in our
discussion at the end of the paper.  These categories partition a full set of
$N$ training data records $D_N$ into $D_{N_A}$ and $D_{N_B}$.  Our idea is to
vary the metadata balance through $N_A$ and $N_B$ in order to explore the the
accuracy of ML models fit to those data. This is represented in the middle  of
Figure \ref{fig:process} as $N_{A_i}$ and $N_{B_i}$ fed into NN 
(an example ML model). Each $(N_{A_i},N_{B_i})$, for $i=1,\dots, r$ is
saved in a vector $x_i^{\top}=(N_{A_i},N_{B_i})$.  Out-of-sample model
accuracy under each $x_i^\top$, denoted as $a_{N_i}$ (e.g.,  correct classification rate 
(CCR) or F1-score), and is saved as $y_i = a_{N_i}$.

The process of varying a metadata balance and recording its corresponding
model accuracy is a computer experiment. Data comprised of $r$ evaluations of
this response surface, which we denote as $(X^{(N)},Y^{(N)})$, can be used to
fit a surrogate model \citep{santner2003design, gramacy2020surrogates} such as
a Gaussian process \citep[GP;][]{williams2006gaussian}. That surrogate can then be searched 
to determine advantageous input conditions $x$ for learning, in this case
determining an optimal metadata balance of future $A$ and $B$ images. This is
shown as the final step of Figure \ref{fig:process}. We refer to this process
as GP-assisted meta-learning (GPAML). We believe we
are the first to apply GP to a performance surface and then view things in a computer 
experiment context.

Although GPAML is a form of learning that leverages a GP, it is fundamentally
different than typical GP active learning (GP-AL) methods. GP-AL methods seek
to either improve \citep{seo2000gaussian,cohn1993neural,mackay1992information}
or optimize \citep{jones1998efficient, snoek2012practical,picheny2016bayesian}
the GP response surface itself, whereas GPAML uses the GP response surface to
improve {\em future} ML model accuracy.  Consequently GPAML is inherently
extrapolative in nature, making it more like ML-AL, but aided by a GP.  We intend
to make this distinction clear as the development progresses. We also note 
that others have used the
word meta-learning in similar contexts; for a review of others, see
\citet{vilalta2002perspective} or \citet{hospedales2021meta}.   Our work is
also distinct from what is known as meta-active learning
\citep[e.g.,][]{hsu2015active, fang2017learning, konyushkova2017learning},
referring to algorithms with flexible acquisition functions that are updated
at each step of the active learning process.

%
%
The remainder of the paper is organized as follows. 
Section \ref{sec:metalearn} contains our main methodological contribution:
GPAML, after a brief review of GPs. Section \ref{sec:results} provides
benchmarking results on three datasets, including the motivating RarePlanes
dataset. Section \ref{sec:discussion} contains a short discussion of
limitations and possible future work. All code is provided via Git:
\url{https://bitbucket.org/gramacylab/metalearn}.

\section{GP-Assisted Meta-learning}\label{sec:metalearn}

We begin with a brief review of GP surrogate modeling and then turn to 
our main contribution.

\subsection{Gaussian Processes}\label{sec:review}

Consider modeling $r$ runs of a computer experiment using a GP
 \citep{sacks1989design}. Let $X=(x_1^{\top}, \dots ,x_r^{\top})$ be an $r
 \times p$ matrix of inputs, and $Y = (y_1, \dots, y_r)$ be a corresponding $r
 \times 1$ vector of outputs. A GP prior implies $Y \sim \mathcal{N}_r
 \bigl(\mu(X), \Sigma(X)\bigr)$, often with $\mu(X)=0$ after centering. This
 induces the likelihood
\begin{align*}
L(Y \mid X ) &\propto \bigl\vert\Sigma(X) \bigr\vert^{-\frac{1}{2}}
 \exp \left(-\frac{1}{2}Y^{\top} \Sigma(X)^{-1} Y\right) 
 \\
 \mbox{where } \quad
\Sigma(X)_{ij}& = \Sigma(x_i,x_j)= \tau^2 \left(\exp \left( - \frac{\left\Vert x_i - x_j 
\right\Vert^2}{\theta} \right) +g\mathbb{I}_{\{i=j\}} \right).
\end{align*}
Although other  {\em kernels} defining $\Sigma(X)$ are common
\citep[e.g.,][]{abrahamsen1997review, stein1999interpolation}, we prefer a
simple squared exponential form.   Hyperparameters
$\tau^2$, $\theta$, and $g$ represent the scale, lengthscale, and noise,
respectively, and may be inferred via maximum likelihood (MLE).

Conditional on data $(X,Y)$ and hyperparameters $\tau^2$, $\theta$, and $g$,
predictions can be made for an $r' \times p$ matrix $\mathcal{X} =
(\tilde{x}_1^{\top}, \dots, \tilde{x}_{r'}^{\top})$ of new inputs. Gaussian
conditioning gives $Y(\mathcal{X}) \mid X,Y \sim
\mathcal{N}_{r'}(\mu_ {\mathcal{X}},\Sigma_{\mathcal{X}})$, where
\begin{align*}
\mu_{\mathcal{X}} & = \Sigma(\mathcal{X}, X) \Sigma(X)^{-1} Y, & \mbox{using } \; \Sigma(\mathcal{X},X)_{ij} &=\Sigma(\tilde{x} _i, x_j), \\
\mbox{and } \quad \Sigma_{\mathcal{X}} & = \Sigma(\mathcal{X}) - \Sigma(\mathcal{X}, X) \Sigma(X)^{-1} \Sigma(\mathcal{X},  X)^\top.
\end{align*}
As shorthand, we use $\mathrm{GP}(\mathcal{X};X,Y)$ to refer to predictions
made by a GP at $\mathcal{X}$ following these equations using MLE
hyperparameters. Deriving such a surface represents the final step of GPAML;
refer to Figure \ref{fig:process}. Since it is trained on model accuracy for
varying metadata composition, we shall refer to $\mathrm{GP}(\mathcal{X};X,Y)$
as an ``accuracy surface'' in this paper.

\subsection{An experiment varying the metadata balance}\label{sec:exp}

Consider a collection of $N$ pairs of data points and their labels $D_N =
\bigl((d_1,\ell_1),\dots, (d_N,\ell_N)\bigr)$ and a holdout set $H_{N'}=\bigl((d'_1,
\ell'_1), \dots (d'_{N'},\ell'_{N'})\bigr)$ of $N'$ new data points and their
labels. In the case of image classification, $d_i$ is the pixel array for the
$i^{\mathrm{th}}$ image and $\ell_i$ is the corresponding classification
label. Suppose that $D_N$ and $H_{N'}$ have been used to train and evaluate,
respectively, a ML model $\mathrm{ML}(H_{N'};D_N)$, e.g., as described in
Section \ref{sec:intro}, producing model accuracy $a_N$. Our goal is to use
$D_N$, but not $H_{N'}$, to decide on what $n$ new data points and labels to
add to $D_N$, and then re-train/evaluate $\mathrm{ML}(H_{N'};D_{N+n})$.
Because both collecting data points and then labeling them is expensive, we
want to ensure that the $n$ new data points collected are useful to the model.
Useful is defined as data which produce the best values of
$a_{N+n}$, the new model accuracy. We exploit metadata information to learn
optimally useful new data points. 
Since we consider only two metadata categories in this work, $A$ and $B$
partitioning $D_N$, choosing $n$ new points is equivalent to determining
the proportion or amount belonging to either category, $n_A$ or $n_B = n - n_A$.

Now, consider an experiment designed to create many different sub-datasets
sampled from $D_N$ via $D_{N_A}$ and $D_{N_B}$, each with different metadata
balances for training, along with hold-out sets for testing so that
accuracy can be measured for each metadata balance. Focus on a single
repetition or instance $i$ of this experiment. Begin by sampling a random
number of images from each metadata category. More explicitly, sample two
random integers $N_{A_i}$ and $N_{B_i}$ representing the number of data points
to select from each category in rep $i$. Ideally, $N_{A_i}$ is a random
integer between 1 and $N_A$, and $N_{B_i}$ is a random integer between 1 and
$N_B$. In practice, the number of points available for training is less than
$N_A$ and $N_B$ because some points must be reserved testing; details on how
the test set is selected are provided in the next few paragraphs. Once the
values of $N_{A_i}$ and $N_{B_i}$ have been selected, use those values to
sample datasets $D_{N_{A_i}} = \mathrm{samp}(D_{N_A}, N_{A_i})$ and
$D_{N_{B_i}} = \mathrm{samp}(D_{N_B}, N_{B_i})$, where $\mathrm{samp}(D,N)$
indicates randomly selecting $N$ images from dataset $D$. Combine these
into $D_{N_i} = (D_{N_{A_i}}, D_{N_{B_i}})$, and use $D_{N_i}$ to train the ML
model.

After selection and training with $D_{N_i}$, we must evaluate the performance
of the ML model out-of-sample. That is, we must compute accuracy
$a_{N_i}$ for some test set $D_{N_i}^{\mathrm{test}}$. The decision of how to
compose $D_{N_i}^\mathrm{test}$ is nontrivial because model accuracy $a_{N_i}$
changes depending on $D_{N_i}^\mathrm{test}$, affecting our accuracy response
surface that ultimately chooses the optimal balance of new test points
$(n_A,n_B)$. More on this in Section \ref{sec:al}. In particular, if
$D_{N_i}^{\mathrm{test}}$ is not similar, in terms of proportion balance in
each metadata category $A$ and $B$, to the true proportion balance, any
downstream decision making may not generalize well.

Here we assume to know, or that we can estimate, the population proportion
balance, denoted $(p_A^0,p_B^0)$. This guides our choice of random testing set,
separately for each rep $i$.  Somewhat arbitrarily, but we think also
sensibly, we choose 10\% of $D_N$ for $D_{N_i}^\mathrm{test}$, meaning the
metadata (count) balance of $D_{N_i}^\mathrm{test}$ is
$(0.1Np_A^0,0.1Np_B^0)$. This restricts the upper bounds for $N_{A_i}$ and
$N_{B_i}$ for training.  Specifically, to ensure $D_{N_i} \cap
D_{N_i}^\mathrm{test} = \emptyset$, we sample $N_{A_i}$ between 1 and
$N-0.1Np_A^0$, and $N_{B_i}$ between 1 and $N-0.1Np_B^0$. Given
$D_{N_i}^\mathrm{test}$, and $D_{N_i} = (D_{N_{A_i}}, D_{N_{B_i}})$ so
constructed, and out of sample accuracy $y_i = a_{N_i}$ associated with
metadata balance $x_i^\top = (N_{A_i}, N_{B_i})$ estimated, the subsampling
process may be repeated for $i=1,\dots,r$ to get
$X^{(N)}=(x_1^{\top},\dots,x_r^{\top})$ and $Y^{(N)}=(y_1,\dots,y_r)$. The
$(N)$ superscript denotes that sub-samples were collected from $D_N$. We will
use $(X^{(N)},Y^{(N)})$ in Section \ref{sec:al} to train a GP and produce an
accuracy surface.

For each unique metadata balance pair, there are many subsamples of data
points that could be selected from $D_{N_A}$ and $D_{N_B}$. In particular, for
a given $N_A$, $N_B$, $N_{A_i}$, and $N_{B_i}$, there are
$\binom{N_A}{N_{A_i}}\binom{N_B}{N_{B_i}}$ possible combinations. Each
subsample will yield a different model accuracy. Therefore, we do not sample
$r$ unique metadata balance pairs, but rather choose to {\em block} the
experiment by a smaller number $b$ unique metadata balance pairs, along with
$z$ ``replicates'' so that $r=bz$.  This helps with separating signal from
noise when fitting the GP surrogate; more in Section \ref{sec:al}. Algorithm
\ref{alg:metavary} summarizes this subsampling process.

\begin{algorithm}[ht!]
\caption{Experiment varying the metadata balance}
\label{alg:metavary}
\begin{algorithmic}[1]

\Require{$N,N_A,N_B ,p_A^0, p_B^0, b, z$}

\For{$j \in 1, \dots, b$}
\State{$N_{A_j} \sim \mathcal{U}(1, N_A - 0.1Np_A^0)$} 
\Comment{sample counts for training set}
\State{$N_{B_j} \sim \mathcal{U}(1, N_B - 0.1Np_B^0)$}
\For{$k \in 1, \dots, z$}
\State{$D_{N_{A_{jk}}}^\mathrm{test} = \mathrm{samp}(D_{N_A},0.1Np_A^0)$} 
\Comment{collect test sets}
\State{$D_{N_{B_{jk}}}^\mathrm{test} = \mathrm{samp}(D_{N_B},0.1Np_B^0)$}

\State{$D_{N_{A_{jk}}} = \mathrm{samp}(D_{N_A} \setminus D_{N_{A_{jk}}}^\mathrm{test}, N_{A_{j}})$}
\Comment{collect training sets}
\State{$D_{N_{B_{jk}}} = \mathrm{samp}(D_{N_B} \setminus D_{N_{B_{jk}}}^\mathrm{test}, N_{B_{j}})$}

\State{$x_{b(j-1)+k} = (N_{A_j}, N_{B_j}); y_{b(j-1)+k}=\mathrm{ML}(D_{N_{jk}}^\mathrm{test}; D_{N_{jk}})$}
\Comment{run ML model and save performance}
\EndFor
\EndFor
\State{$X^{(N)}=(x_1^{\top}, \dots, x_{bz}^{\top})$}
\Comment{combine data from all iterations}
\State{$Y^{(N)} = (y_1, \dots, y_{bz})$}
\end{algorithmic}
\end{algorithm}

\noindent 
$\mathcal{U}(a,b)$ in lines 2 and 3 refers to a sample from a discrete uniform
distribution over $\{a, a+1, \dots, b\}$.
 
 \subsection{Conic meta-learning}\label{sec:cone}
 
We now present an idea for meta-learning, called conic meta-learning, that can
be used for any accuracy (response) surface.  We first use an arbitrary surface to 
describe the meta-learning approach for adaptively adding data, ignoring the ML 
model and corresponding surrogate fit to simplify the experiment. 
We return to actual meta-learning
with a GP in Section  \ref{sec:al}. For now, imagine an accuracy surface
trained on data $(X^{(N)},Y^{(N)})$, where $X^{(N)}$ and $Y^{(N)}$ have the
same meaning as in Section \ref{sec:exp}, except that they follow
\begin{equation} y_i \sim \mathcal{N}(f(x_1,x_2), 0.05^2), \quad 
\mbox{where} \quad f(x_1,x_2)=
1- \exp \Bigl(-\frac{x_1  x_2}{10x_1 + 15x_2}\Bigr), \label{eq:toy}
\end{equation}
and $x_1$ and $x_2$ represent the number of data points subsampled from
category $A$ and category $B$, respectively (i.e., values of $N_{A_i}$ and
$N_{B_i}$). Data simulated from the process (\ref{eq:toy}) have important qualities as
relates to our actual meta-learning application: (i) the mean function is
monotonic in each dimension and (ii) the response $y_i$ is noisy. Monotonicity
is important because ML models generally get more accurate as more data is
added; noise is important since controlling metadata does not pin down the
actual training/testing examples in the partition, which is our reason for
blocking in Algorithm \ref{alg:metavary}.

\begin{figure}[ht!]
\centering 
\includegraphics[scale=0.44, trim = {20, 15, 50, 20}, clip]{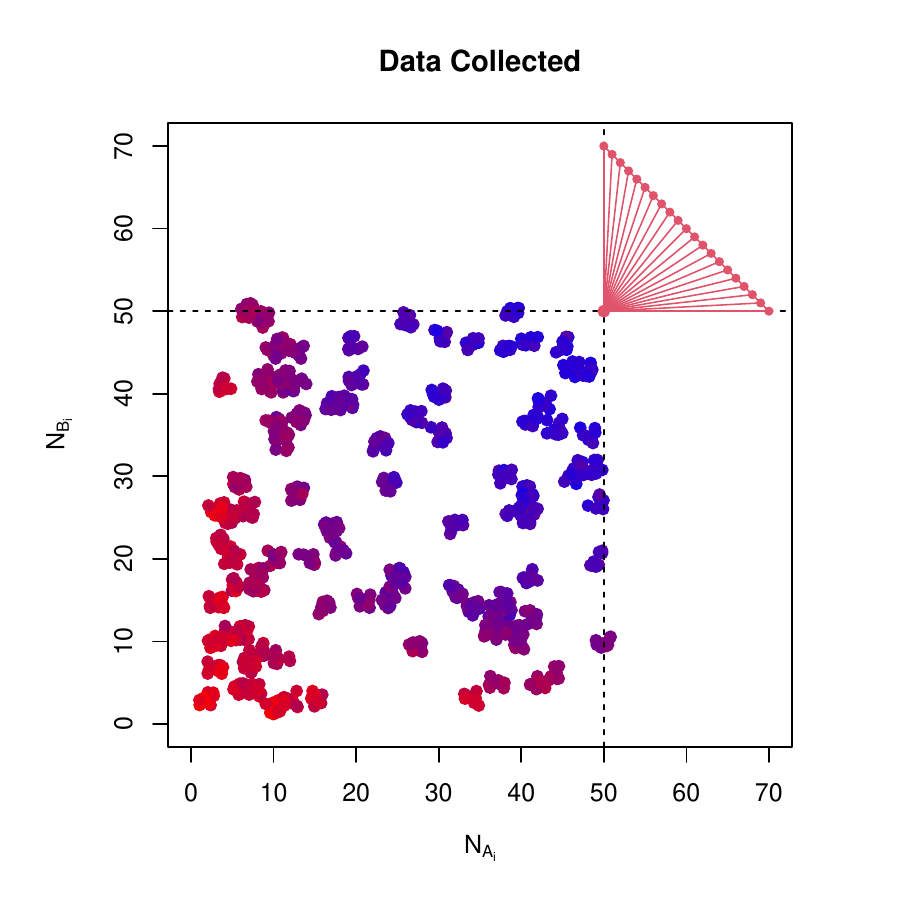}
\hfill
\includegraphics[scale = 0.44, trim = {40, 15, 50, 20}, clip]{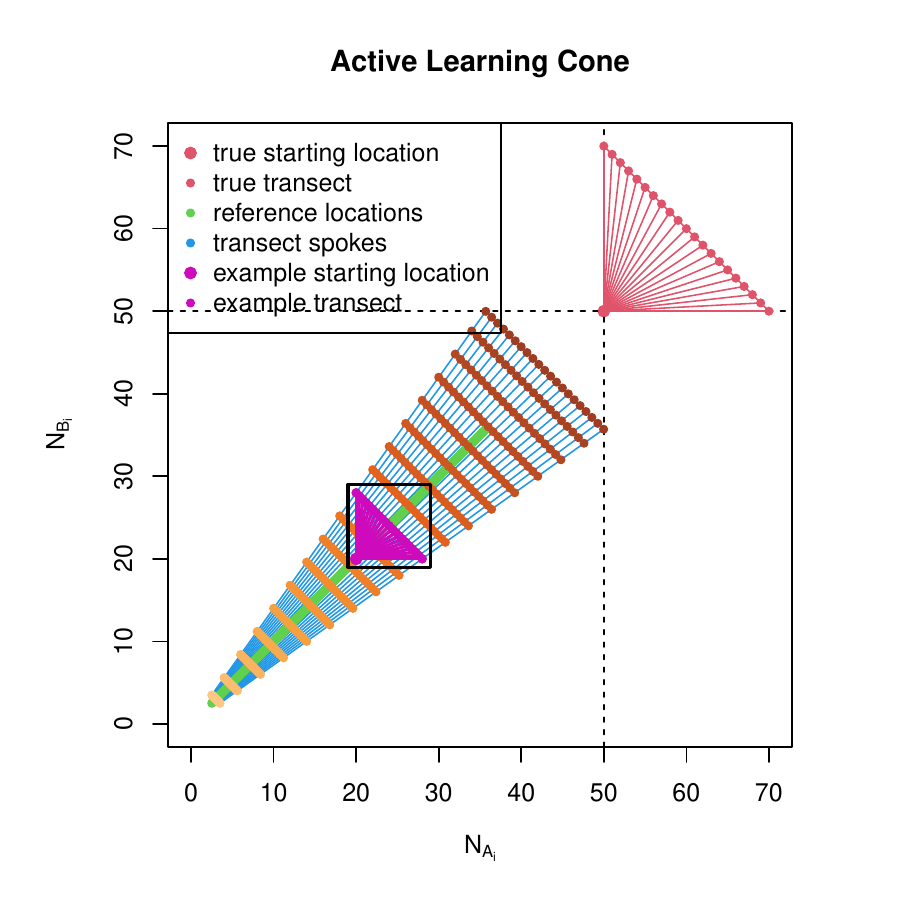}
\hfill
\includegraphics[scale = 0.44, trim = {40, 15, 50, 20}, clip]{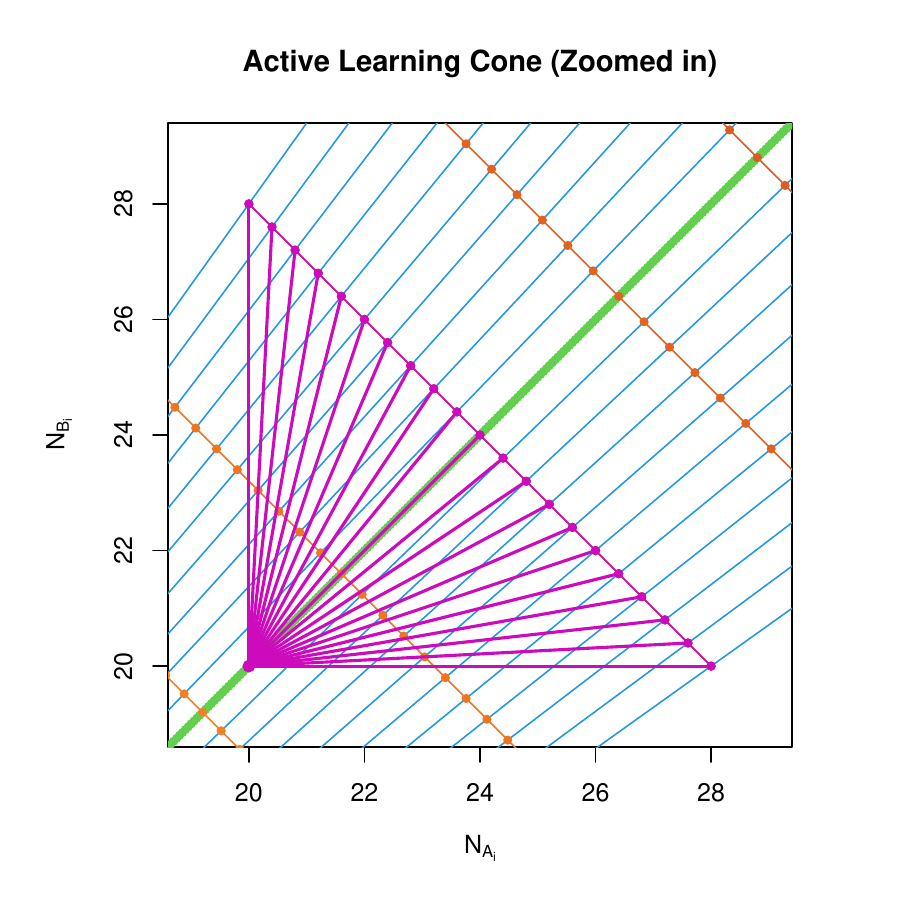}
\caption{{\em Left:} Sample data points $X^{(N)}$, colored by value of $Y^{(N)}$. 
{\em Center:} Cone resulting from all possible reference locations. {\em Right:} 
Center plot, zoomed in to boxed region. This shows all possible paths for the reference 
location $(20,20)$.}
\label{fig:cone}
\end{figure}

The left panel of Figure \ref{fig:cone} provides an example/representative
dataset following Eq.~(\ref{eq:toy}). The $x$- and $y$-coordinates on the plot
represent $x_1$ and $x_2$, or metadata categories $A$ and $B$, and the color of
each dot indicates the level of $y$. Colors closer to blue have higher model
accuracy, and colors closer to red have lower model accuracy. Clusters of
points on the plot come from replicates in the blocked design, shown with
jitter to see all values of $y$ in the block. For this example we assume
$N=100$, $N_A=N_B=50$, and that we wish to add $n=20$ new points. The true
starting metadata balance in $D_N$ is $(N_A, N_B) = (50,50)$ and is indicated as
a red dot.
 
When starting with a balance of $(50,50)$ and adding $n=20$ new points, there
are exactly 21 ways new points $(n_A, n_B)$ can be added with respect to the
metadata category balances: $\bigl((0,20),(1,19), \dots, (20,0)\bigr)$. By
adding each of these 21 possible new point configurations to our current
metadata balance $(50,50)$, we get a {\em transect matrix}  $T=\bigl((50,70)^{\top},
(51,69)^{\top}, \dots, (70, 50)^{\top}\bigr)$ of new possible metadata
balances. 
In the middle panel of the figure, lines connecting the starting balance
$(50,50)$ to each point of the transect are shown in red, representing the 21
possible ``paths'' of meta-learning. Our goal is to choose the path which
provides the best model accuracy.

To choose the best meta-learning path, we use the accuracy surface created by
training a model on $(X^{(N)},Y^{(N)})$. Notice in the left two panels of
Figure \ref{fig:cone} that all samples in $(X^{(N)},Y^{(N)})$ are bounded by
50 in both dimensions, emphasized using the dashed lines. This is because we
only have 50 data points from each category to choose from (i.e., starting
metadata balance $(50,50)$). Therefore, the predictions in the accuracy
surface are only reliable inside the dashed lines. This is a problem, because
the data points of interest (the transect $T$), lie outside of the bounds of
the model (i.e., we are trying to extrapolate). We need a way to connect the
data points in $T$ to data points in the accuracy surface so that we can use
the accuracy surface to make a decision. Our key assumption in this paper is
that the behavior of the accuracy surface is roughly stationary with regards
to proportion balance. In other words, that we may use data points with the
same proportion balance as $T$, but with smaller $N_{A_i}$ and $N_{B_i}$, to
decide on acquisitions.

Consider representing the metadata balance as a proportion instead of a count.
The true metadata balance $(50,50)$ has a corresponding proportion balance
$(\frac{N_A}{N}=\frac{1}{2}, \frac{N_B}{N}=\frac{1}{2})$. Within the range of
the accuracy surface, we have many data points showing model performance when
the proportion balance is $(\frac{1}{2},\frac{1}{2})$; one example is at
$(20,20)$. If we were to visualize all data points with this proportion
balance, we would have a line at $y=\frac{N_B}{N_A}x$. This line is shown 
in green in middle and right panels of
Figure \ref{fig:cone}. We refer to the points similar to $(50,50)$ as
reference locations $R_q = \mathrm{ref}(\frac{N_A}{N},\frac{N_B}{N}, q)$,
where $q$ is the number of reference locations. The notation
$\mathrm{ref}(\frac{N_A}{N},\frac{N_B}{N}, q)$ represents a matrix of $q$
reference locations with the proportion balance
$(\frac{N_A}{N},\frac{N_B}{N})$, where each reference location is saved in a
separate row. Our stationarity assumption means that model performance of the
reference locations are representative of performance near $(N_A,N_B)$, and
beyond for $(N_A + n_A, N_B + n_b)$.

Now that we have points similar to the true metadata balance, we wish to
parlay that into transects similar to $T$. In other words, we want to find the
transect $T_j^\mathrm{ref}$ corresponding to each reference location $R_{j}$,
say in a grid indexed by $j=1,\dots,q$. To show how to find these transects,
consider a single reference location $(20,20)$. This location is shown in pink
in the right two panels of the figure; the right panel is zoomed in on the box
highlighted in the center panel. To find a transect with respect to $(20,20)$
that is representative of the transect with respect to $(50,50)$, consider
starting at $(20,20)$ and creating paths that are ``equivalent'' to the paths
starting at $(50,50)$. Equivalent paths are defined as those which have the
same percentage change in points as the true location. In our example, we are
adding $n=20$ new points to a dataset of size $N=100$, which is a 20\%
increase in the size of the dataset. To add an equivalent number of points to
the $(20,20)$ reference location, we would add $0.2\times40=8$ new points.
Adding an equivalent number of points can yield fractional points along the
transect, but this is okay since a surrogate would approximate the entire
response surface, not just integer inputs.  For the $(20,20)$ reference
location, the transect would be $T_i^\mathrm{ref}=\bigl((20,28)^{\top}, (20.2,
27.8)^{\top}, \dots, (28,20)^{\top}\bigr)$.  By adding new points according
the percentage increase in size, the 21 points along each reference location's
transect have the same ending proportion balances as the original transect
$T$.

The same process can be repeated for a dense sequence of reference locations
(the green line in Figure \ref{fig:cone}). The complete array $T^\mathrm{ref}$
is a 3D array of length $j=1,\dots,q$, where each entry is a $21 \times 2$
transect matrix. Lines connecting the 21 points of each transect are shown in
blue.  There are also many transects which are highlighted in shades of
orange; these will be used to connect the visual to other plots in the next
subsection where we shall look at the value of the accuracy surface at all of
the points in all of the transects. Because we want to ensure these
predictions are valid, we ensure that all transects are bounded by the model.
We do this by only considering reference locations corresponding to transects
that are bounded by the dashed lines. In particular, notice that the blue
lines of the cone are all bounded by the dashed lines. The collection of all
transects is conic in structure when plotted.  For this reason, we refer to
meta-learning done using this process as conic meta-learning.

\subsection{Learning the optimal metadata balance}\label{sec:al}

\begin{figure}[ht!] 
\includegraphics[scale=0.42, trim = {20, 15, 40, 20}, clip]{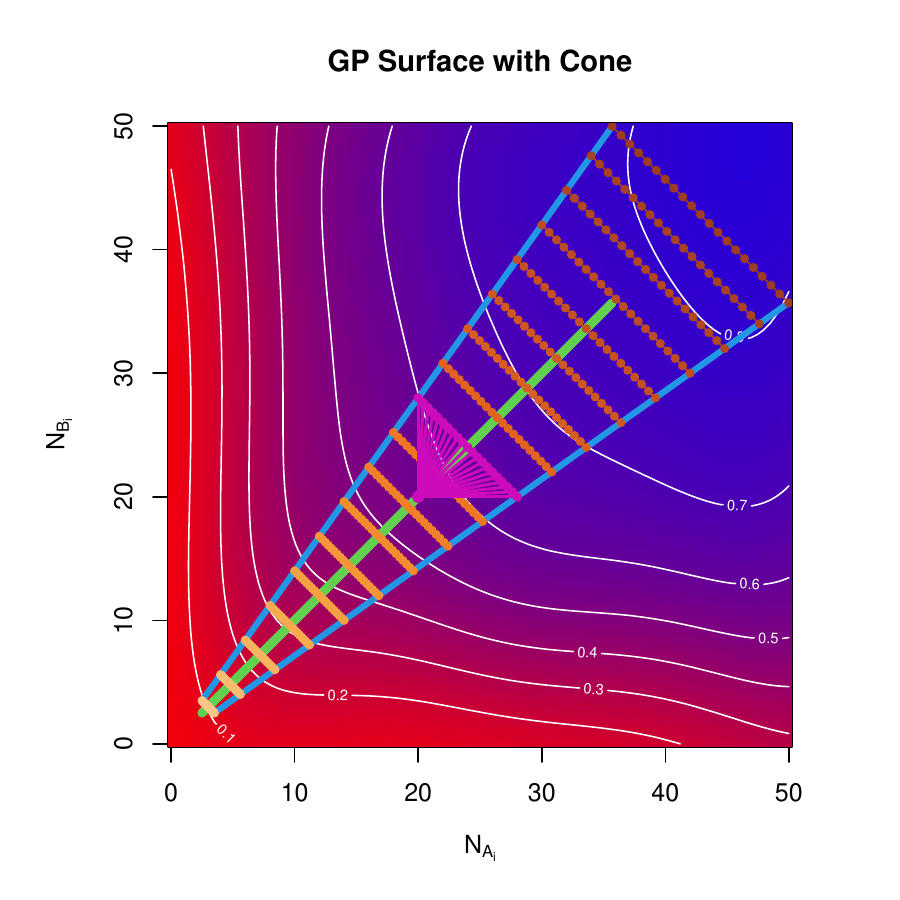}
\hfill
\includegraphics[scale=0.42, trim = {25, 15, 5, 20}, clip]{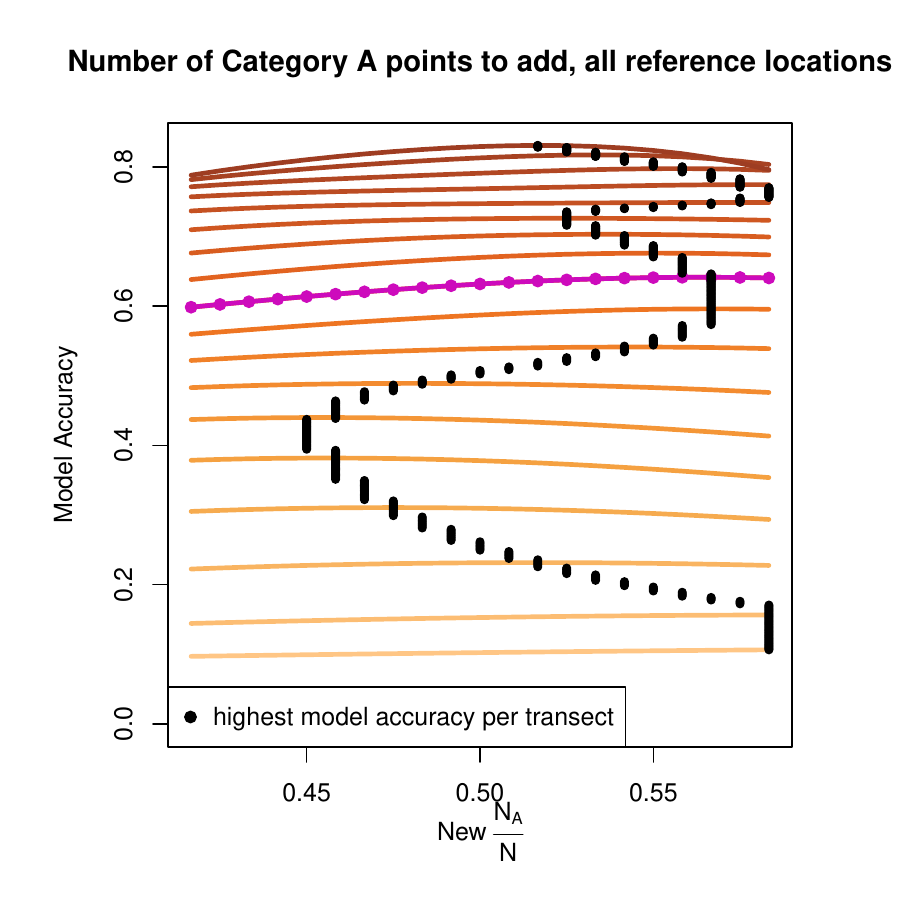}
\hfill
\includegraphics[scale=0.42, trim = {25, 15, 40, 20}, clip]{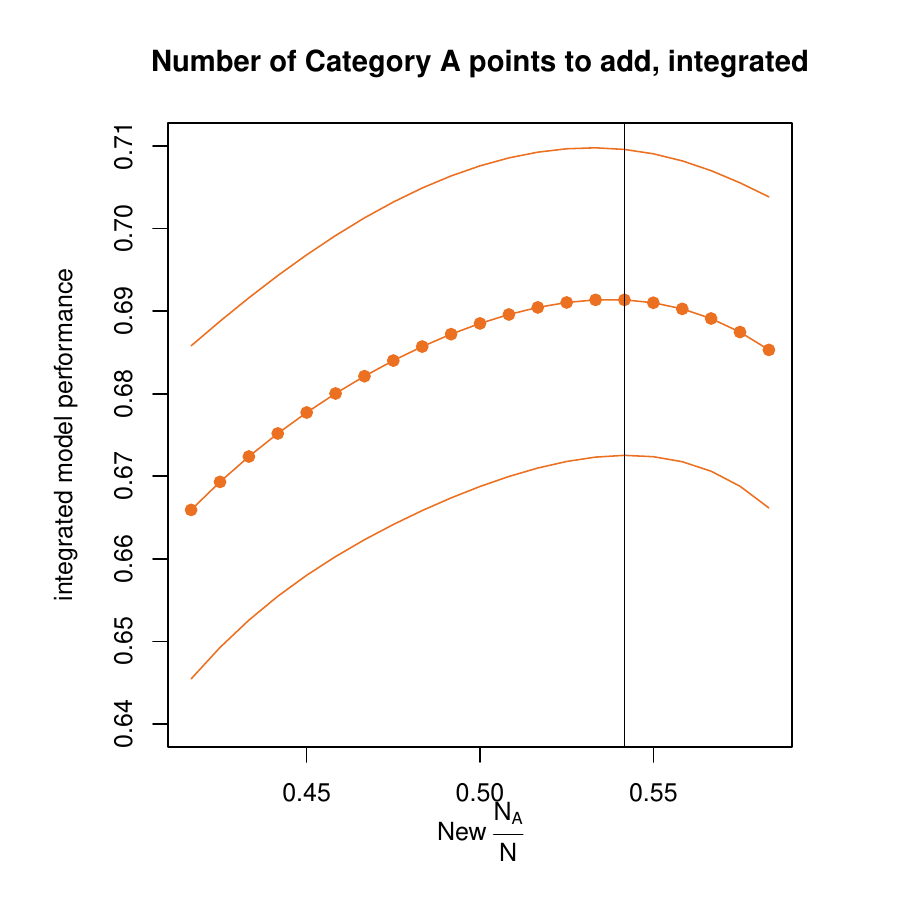}
\caption{{\em Left:} GP surface with cone overlaid. {\em Center:} 
Transects of cone plotted against model performance. {\em Right:} 
Integrated model performance.}
\label{fig:gpcone}
\end{figure}

Now that we have a collection of transects $T^\mathrm{ref}$ corresponding to a
collection of reference locations $R_q$, we want to combine the transects in
such a way that we find the single optimal metadata balance. To do that, we
must incorporate a surrogate model into our example problem. Using the
example data we explored in Section \ref{sec:cone}, we fit a GP trained on
data $(X^{(N)},Y^{(N)})$ to get the accuracy surface. The accuracy surface we
are using is shown in the left panel of Figure \ref{fig:gpcone} with the cone
constructed in Section \ref{sec:cone} overlaid.
Only the outside edges of the cone are shown for visual simplicity. In this plot, 
colors represent predicted accuracy at each location; locations that 
are more red have lower model accuracy and locations that are more blue 
have higher model accuracy. The contours in white show the same information. 
Notice again that the mean surface is monotonic in each dimension, since the model 
performs better on average as more data is added. The predicted accuracy  
 at each input combination will be important for understanding the next step of 
the meta-learning process.

We now use the GP to predict the model accuracy for all transects,
$\mathrm{GP}(T^\mathrm{ref}; X^{(N)},Y^{(N)})$, saving the mean prediction in
$\mathcal{Y}$. Recall that $T^\mathrm{ref}$ is an array, with each entry
$T^\mathrm{ref}_i$ comprising of a single $21 \times 2$ transect matrix.
Predictions are made at each entry $T^\mathrm{ref}_i$ separately, and for each
entry the mean prediction is a vector of length 21. Doing this for all
transects $T^\mathrm{ref}$ produces $q$ vectors of length 21, which are saved
in a $21 \times q$ matrix. In this way, each row of $\mathcal{Y}$ represents a
single ending proportion balance.

The center panel of Figure \ref{fig:gpcone} offers a visual. Orange lines in
this plot correspond to orange lines in the cone in the left panel; i.e, each
line represents a transect. As before, the $(20,20)$ transect is shown in pink for
reference.  The center panel plots the ending proportion balance (with respect
to Category $A$) associated with each of the 21 points in the transect against
the predicted mean accuracy. Each orange line in the plot is a single column
of $\mathcal{Y}$, where the $x$-axis is related to the entry number and the
$y$-axis is the entry of that column. Across all columns, observations with
the same entry number have the same ending proportion balance, indicated on
the $x$-axis of the plot. For each transect, one of the 21 entries provides
the maximum model accuracy. We show the entry with the best model accuracy for
each transect as a black dot. This represents the optimal proportion balance
for that transect. Note that every transect does not suggest the same optimal
proportion balance.

The final step of this meta-learning process is to collapse $\mathcal{Y}$ into
a single piece of information. We know that each column of $\mathcal{Y}$
suggests a single optimal proportion balance, but that the optimal proportion
balance is different for each column. We propose collapsing $\mathcal{Y}$ into
a single column by computing a weighted average over the columns.
In particular, we compute $G = \mathcal{Y}w$  to get a
single vector $G$ with 21 entries. Weights $w$ are chosen such that more
weight is given to transects in the upper right hand corner of the plot,
because they are the most similar to the true transect $T$. 
We use linear weighting, but other approaches could be used to adjust  
the impact of transects further from the starting balance (i.e., the transects 
closer to the bottom left corner of the accuracy surface). The resulting vector $G$ is plotted in the right panel of Figure
\ref{fig:gpcone}, with respect to Category $A$. The proportion balance
associated with the entry number of $G$ is on the $x$-axis, and the entry of
$G$ (the averaged model accuracy) is on the $y$-axis.

The proportion balance that produces the best value of $G$ serves as our
acquisition choice, i.e., 
$(n_A,n_B) = \arg
\max_T(G)$.
The optimally chosen proportion balance from our running example is shown as a
vertical line in the right panel of Figure \ref{fig:gpcone}. Here, we learn
that the optimal new proportion balance is $(\frac{65}{120},\frac{55}{120})$,
associated with transect entry $(65,55)$, yielding  $(n_A,n_B)=(15,5)$. An
algorithm detailing the steps explained in Sections \ref{sec:cone} and
\ref{sec:al} is shown in Algorithm \ref{alg:cone}, shorthanded as GPAML.
 \begin{algorithm}[ht!]
\caption{GPAML}
\label{alg:cone}
\begin{algorithmic}[1]

\Require{$X^{(N)}, Y^{(N)}, N,N_A, N_B, n, f, w$}
\State{Complete Algorithm \ref{alg:metavary}.}
\State{$R_q=\mathrm{ref}(\frac{N_A}{N},\frac{N_B}{N}, q)$}
\Comment{create a sequence of $q$ reference locations}
\State{$T = \bigl( (N_A + n, N_B)^{\top}, (N_A+n-1,N_B+1)^{\top}, \dots, (N_A, N_B+n)^{\top} \bigr)$}
\Comment{create transect matrix}
\For{$i \in 1, \dots, q$}
\State{$n_\mathrm{ref} = \frac{\sum R_{q_i}}{N}$}
\Comment{find equivalent move size for reference location}
\State{$T^\mathrm{ref}_i = n_\mathrm{ref}T$}
\Comment{get transect for reference location $i$}
\EndFor
\State{$\mathcal{Y} = \mathrm{GP}(T^\mathrm{ref}; X^{(N)},Y^{(N)})$ }
\Comment{make GP predictions at reference location transects}
\State{$G= \mathcal{Y}w$}
\Comment{take weighted average of GP predictions}
\State{$(n_A,n_B)=\arg\max_T(G)$}
\Comment{choose $(n_A,n_B)$ which optimizes $G$}
\end{algorithmic}
\end{algorithm}

\begin{figure}[ht!]
\includegraphics[scale = 0.37]{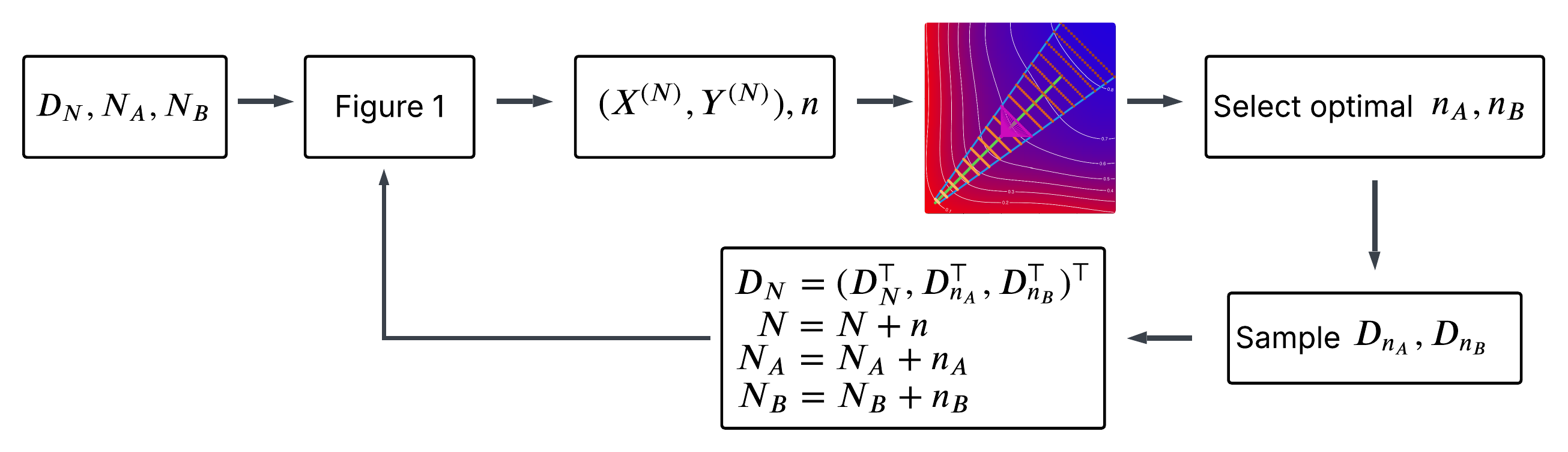}
\caption{Multi-step GPAML, iterating over data sizes $N=N_{\mathrm{start}},\dots,
N_{\mathrm{stop}}$.}
\label{fig:gpal}
\end{figure}

GPAML can be completed many times in a row to make many consecutive batches of
data acquisition, starting at data size $N_\mathrm{start}$ and ending at data size
$N_\mathrm{stop}$. We present a graphic in Figure \ref{fig:gpal}
showing what this looks like in the context of GPAML.
In this plot, the box labeled ``Figure 1'' refers to the process shown in
Figure \ref{fig:process} up to the creation of $(X^{(N)}, Y^{(N)})$. The next
panel, the GP surface with cone overlaid, represents using a GP surface to
perform conic meta-learning. After using GPAML to select the optimal balance
$(n_A,n_B)$ and collect new data points accordingly, they are appended to the
existing data. The new dataset $D_N$ can be used for data subsampling, and the
whole process can be repeated indefinitely. This allows researchers to
continue collecting data until ``enough'' data has been collected. The
question of when to stop collecting data is determined by a given problem's 
budget and other constraints. Our multi-step GPAML approach guides data 
collection choices for the allocated budget.

\section{Implementation and results}\label{sec:results}

Here we present three examples to demonstrate the effectiveness of GPAML. 
A summary of each dataset/experiment is included in Table \ref{tab:exp}, with further details 
in subsections to follow. 
\begin{table}[ht!]
\begin{tabular}{c|c|c|c|c|c|c|c}
\textbf{Dataset} & \textbf{ML Model} &  \boldmath$N_\mathrm{start}$ &  \boldmath$N_\mathrm{stop}$ & \boldmath$n$ & \textbf{Metadata} & \textbf{Reps} & \textbf{Perf. Criterion}\\
\hline
Spambase & RF & 100 & 500 & 20 & Char/No char & 100 & CCR\\
\hline 
MNIST & LeNet-5 & 100 & 200 & 20 & Easy/hard & 100 & CCR\\
\hline
RarePlanes & YOLOv8 & 100 & 1000 & 50 &Snow/Not snow & 3 & F1 Score\\
\end{tabular}
\caption{Implementation information for each dataset/experiment.}
\label{tab:exp}
\end{table}
For each experiment we consider an initial dataset of size $N_\mathrm{start}$,
and add $n$ new data points at a time until reaching size $N_\mathrm{stop}$. 
At each size of the experiment $N$ we evaluate out-of-sample (OOS) model performance with a
holdout set $H_{N'}$ of size $N'=1000$. Model performance is evaluated using
either correct classification rate
$\mathrm{CCR}=\frac{\mathrm{TP}+\mathrm{TN}}{\mathrm{TP}+\mathrm{FN}+\mathrm{TN}+\mathrm{FN}}$
or F1 score
$\mathrm{F1}=\frac{\mathrm{TP}}{\mathrm{TP}+\frac{1}{2}(\mathrm{FP}+\mathrm{FN})}$,
where TP, FN, TN, and FN represent the number of true positives, false
negatives, true negatives, and false negatives, respectively. For each
experiment the GP is trained on $r=1000$ reps at $b=100$ unique metadata
balances. We consider three competing methods for data acquisition in our
benchmarking exercises:
\begin{itemize}
\item \textbf{GPAML}: Our method as described in Section \ref{sec:metalearn}.
\item \textbf{Random}: New data points are selected at random from a repository of data points.
\item \textbf{Random Action}: The new metadata balance $(n_A,n_B)$ is chosen at random.
\end{itemize}
All experiments were hosted in \textsf{R}, with NN models run in \textsf{python}. 
Code is provided at \url{https://bitbucket.org/gramacylab/metalearn}.
We use the
 \texttt{RcppAlgos} \citep{rcpp}  package to generate the transect vector $T$, and 
 the \texttt{mleHomGP()} function in the \texttt{hetGP} \citep{hetGP} package 
with default settings to fit the GP.

\subsection{Spambase}\label{sec:spam}

The first dataset we consider is the Spambase dataset \citep{spam}, which summarizes
a collection of 4,601 emails. It contains a total of 57 input variables
describing features of the emails, and one output variable classifying the
email as spam or non-spam. The input variables measure the frequency of
certain words, numbers, special characters, and capital letters in the email.

There is technically no metadata provided with the dataset, so we engineered
metadata categories for this experiment. We combined variables describing the frequency of
specific special characters into one metadata variable describing the presence
or absence of special characters in an email, and then removed the original
variables as predictors available for training. So we have two metadata
categories: $A$, emails with no special characters; and $B$, emails with at
least one special character. We chose this metadata among other possibilities
because exploratory analysis revealed that changing the distribution of
special character/no special character emails affected ML model performance; 
see Appendix \ref{sup:spammeta}.

For this experiment we started with $N_\mathrm{start}=100$ emails, adding
$n=20$ data points at a time until reaching size $N_\mathrm{stop}=500$. At
each rep of the experiment, we choose a random number between 20 and 80 to
represent the starting number of emails with at least one special character.
This prevents choosing a starting metadata balance that would make GPAML
infeasible (i.e., a very unbalanced dataset; we discuss this further in Section 
\ref{sec:discussion}). We use CCR to measure model
accuracy. We use a random forest (RF) as our ML model and utilize the
\texttt{randomForest} \citep{liaw2002rf} package.
 
\begin{figure}[ht!]
\includegraphics[scale=0.35, trim = {25, 15, 45, 20}, clip]{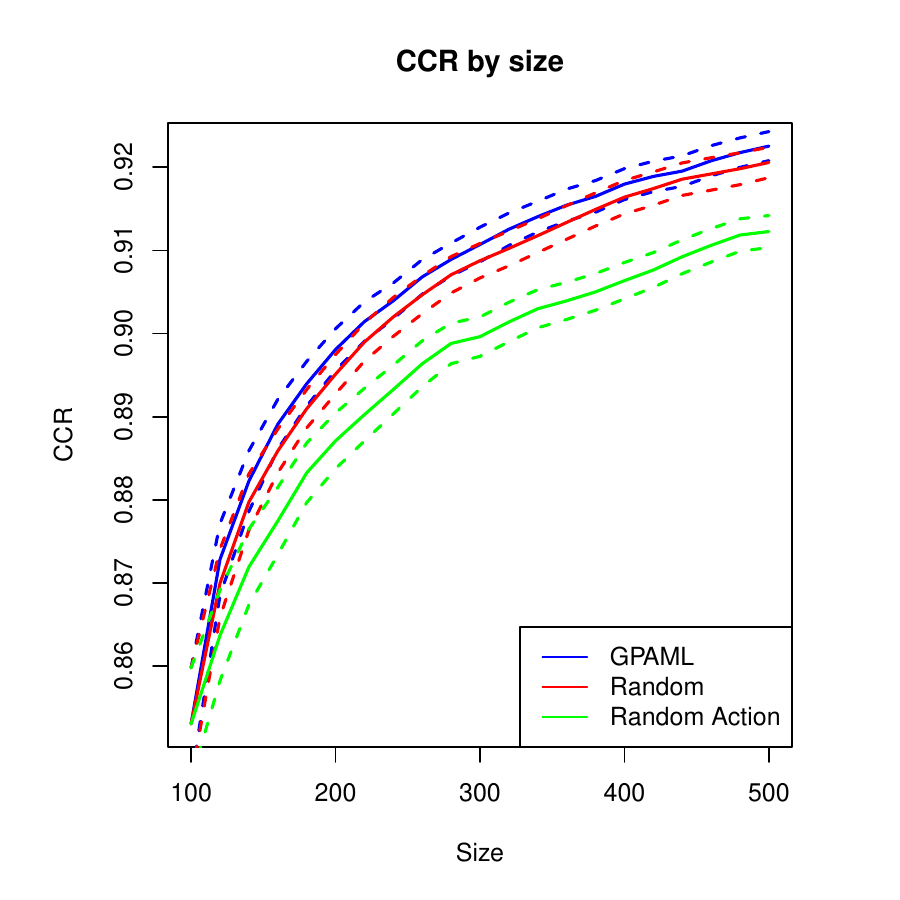}
\hfill
\includegraphics[scale=0.35, trim = {25, 15, 50, 20}, clip]{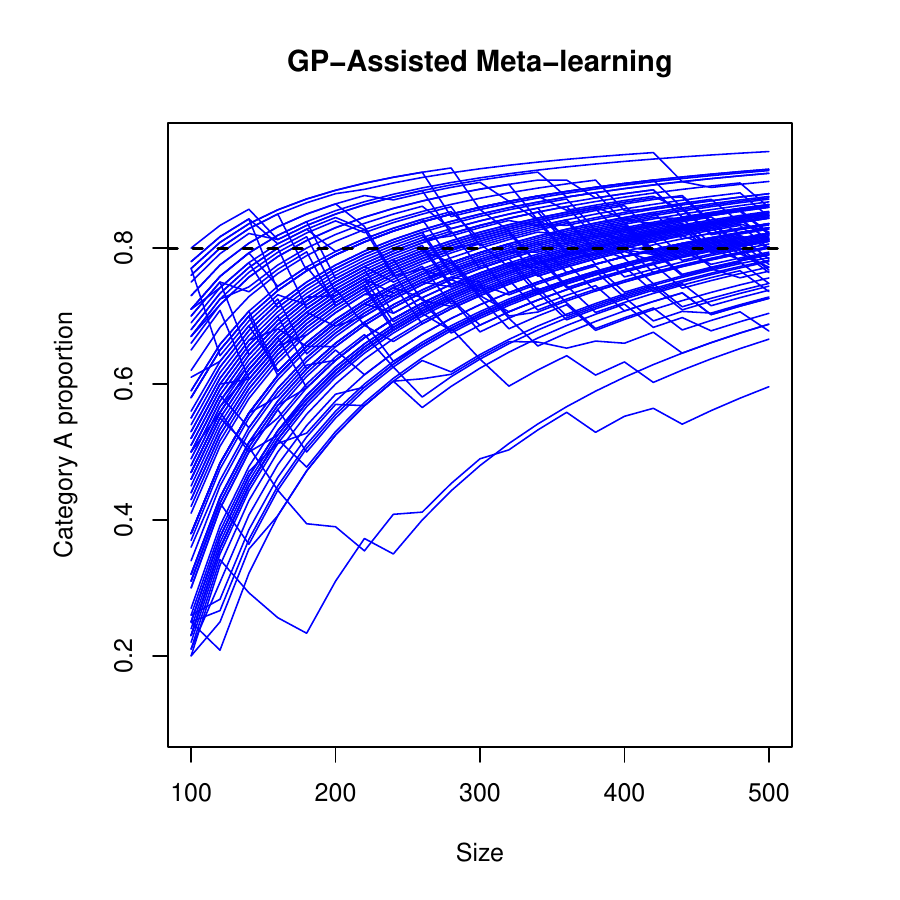}
\hfill
\includegraphics[scale=0.35, trim = {40, 15, 50, 20}, clip]{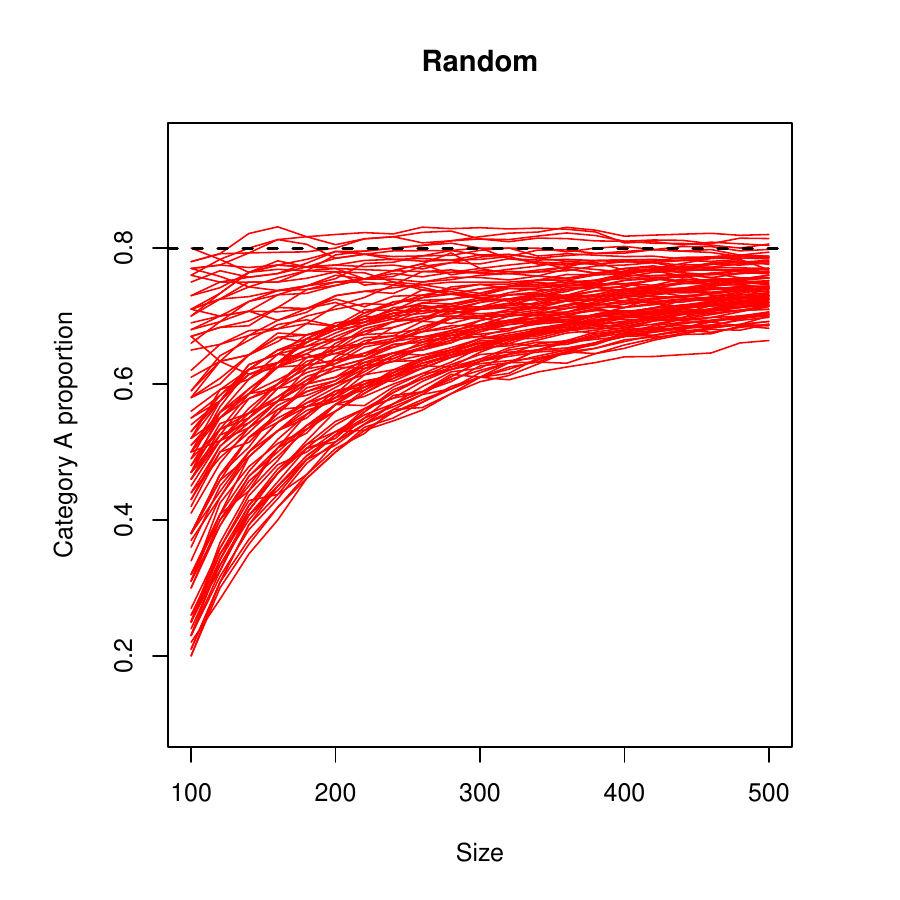}
\hfill
\includegraphics[scale=0.35, trim = {40, 15, 50, 25}, clip]{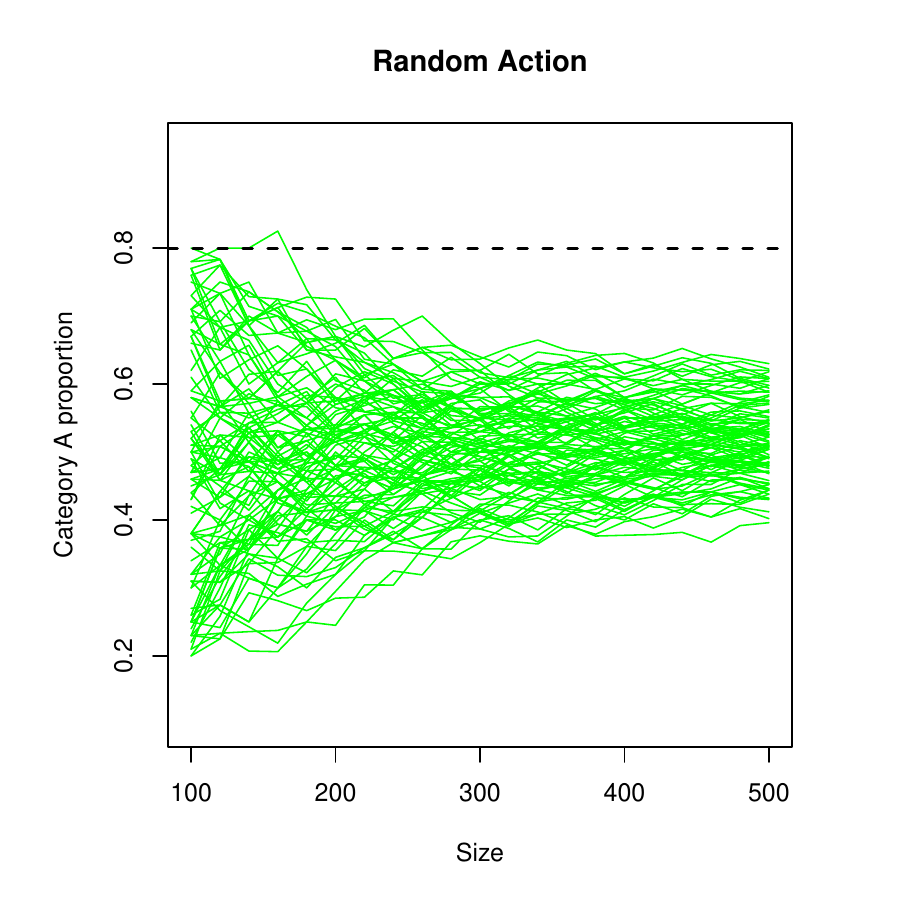}
\caption{{\em Left:} OOS performance for each of the three methods. Proportion balances 
for GPAML {\em (center left)}, random {\em (center right)}, and random action {\em (right)} 
for each rep of experiment.}
\label{fig:spam}
\end{figure}

Out-of-sample CCR, averaged over 100 reps, is shown in the left panel of
 Figure \ref{fig:spam}.
We provide mean CCR and 95\% confidence bounds for each of the three methods.
Observe that GPAML performs as well or better than both of the competitor
methods, in particular over the random action method. Although it does not
perform much better than the random method, that is not necessarily a bad
thing. Sampling new emails completely at random is not a realistic AL approach
since we do not typically have access to the entire repository of data. In
many cases, random selection is approximately optimal; we will discuss this
further in Section \ref{sec:discussion}. What is important is that GPAML has
protected us from making a data acquisition that hurts model performance.

In the right three panels of Figure \ref{fig:spam} we show the proportion
balance for each rep of the experiment for each of the three methods, with
proportion shown with respect to the character category. We also show a dashed
line at $x=0.8$, which is the true balance of special character/no special character emails,
for reference. First notice that the random and random action methods converge
to proportion balances of 0.8 and 0.5, respectively. Then notice that GPAML
also converges roughly to a proportion balance of 0.8, more quickly but with higher
variability than the random method. This indicates that the optimal proportion
balance may vary slightly from one random training/testing partition to
another. This is a strength of GPAML, since other methods of data acquisition
do not consider the optimal proportion balance.

\subsection{MNIST}\label{sec:mnist}

The next dataset we consider is the Modified National Institute of Standards
and Technology (MNIST) handwritten digit database \citep{lecun1998mnist}. This
is a collection of 60,000 images of handwritten digits from 0-9. The dataset
is widely available; we accessed it using the \texttt{dslabs} \citep{dslabs}
\textsf{R} package. An LeNet-5 \citep{lecun1998gradient} NN architecture was
used for the ML model, leveraging the \texttt{PyTorch}
\citep{paszke2019pytorch} package, and CCR was used to evaluate out-of-sample
accuracy. Again, there is no freely available metadata, so we engineer metadata categories.
Based on preprocessing, we binned each of the ten digits into digits that are
easy to categorize and digits that are difficult to categorize. The digits in
the easy, $A$ category are 0, 1, 4, 5, 6, and 7 and the digits in the hard,
$B$ category are 2, 3, 8, and 9; see Appendix \ref{sup:mnistmeta} to see 
how these categories were chosen.

\begin{figure}[ht!]
\includegraphics[scale=0.35, trim = {25, 15, 45, 20}, clip]{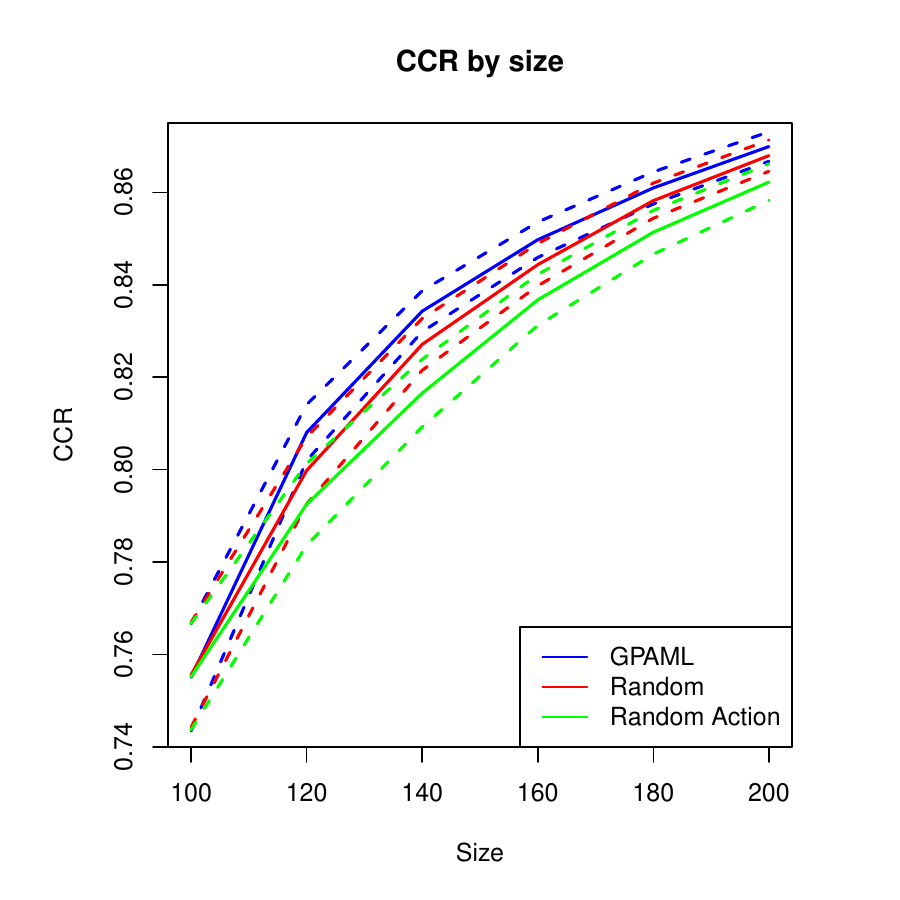}
\hfill
\includegraphics[scale=0.35, trim = {25, 15, 50, 20}, clip]{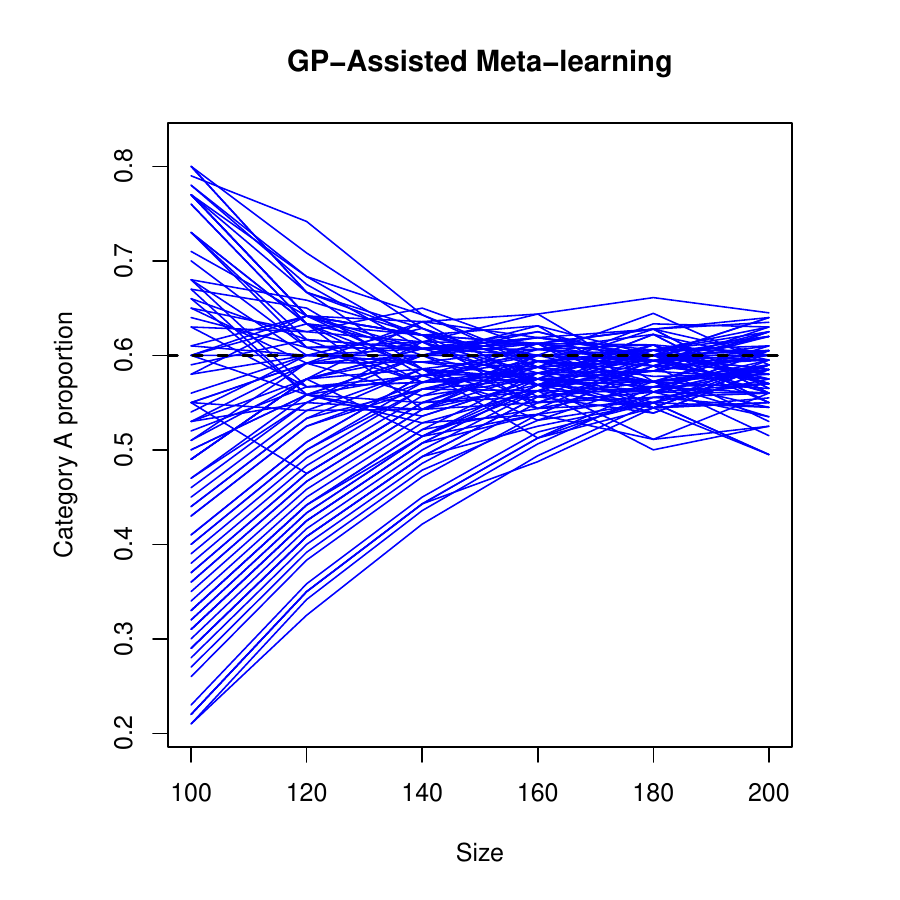}
\hfill
\includegraphics[scale=0.35, trim = {40, 15, 50, 20}, clip]{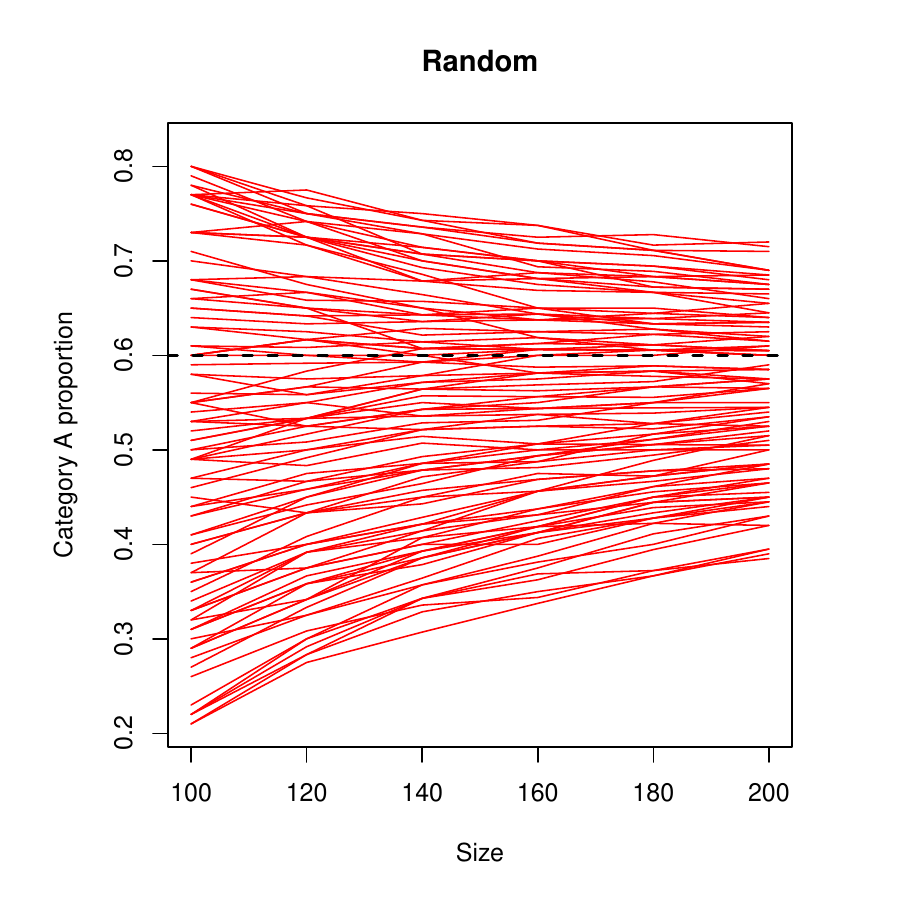}
\hfill
\includegraphics[scale=0.35, trim = {40, 15, 50, 25}, clip]{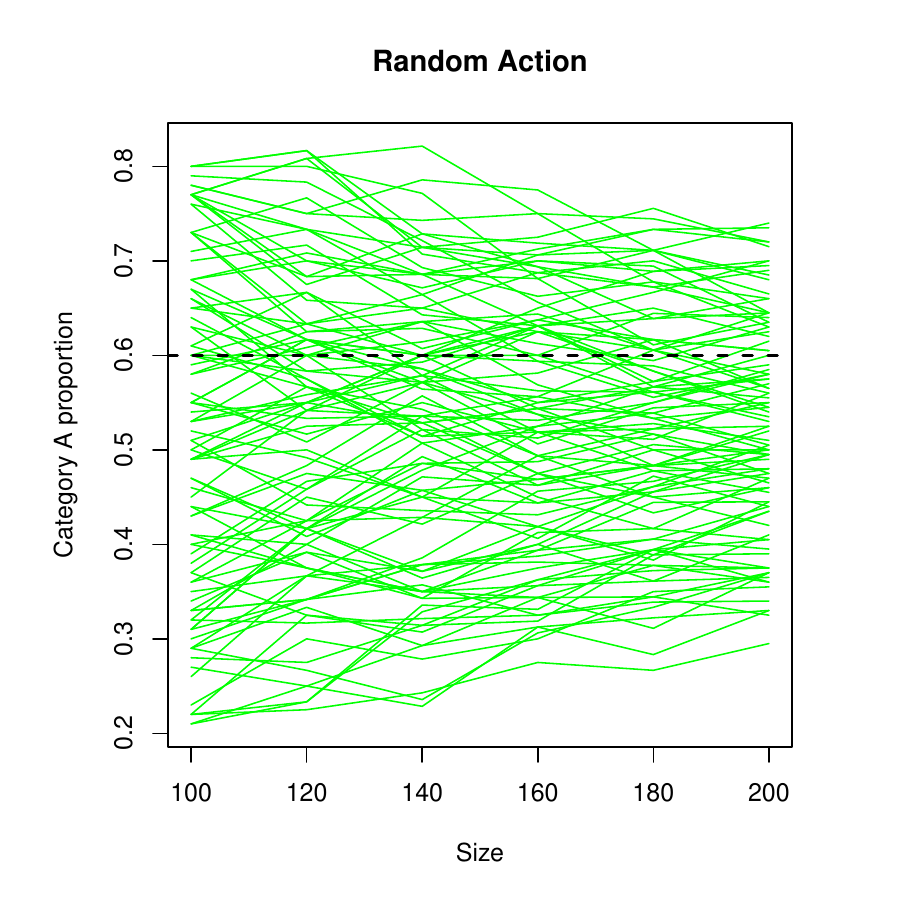}
\caption{{\em Left:} OOS performance for each of the three methods. Proportion balances 
for GPAML {\em (center left)}, random {\em (center right)}, and random action {\em (right)} 
for each rep of experiment.}
\label{fig:mnist}
\end{figure}

For this experiment we began with $N_\mathrm{start}=100$ images, adding $n=20$
new images at a time until reaching size $N_\mathrm{stop}=200$. For a single
rep of the experiment, we choose a random starting number of easy images
between 20 and 80.  Results of this experiment, averaged over 100 reps, are
shown in the left panel of Figure \ref{fig:mnist}
via mean CCR and 95\% confidence bounds for each of the three methods. Again,
notice that GPAML performs as well or better than both of the competitor
methods, particularly random action. This is most evident from sizes $N=120$
to $N=160$ of the experiment; by size $N=200$, random has caught up to GPAML.

The other three panels of the figure show overall proportion balance at each
step of the meta-learning process. The proportion balance of each rep for
GPAML, random, and random action is shown in the right three panels of Figure
\ref{fig:mnist}. The proportion balance is shown with respect to the easy
category. We also include a dashed line at $y=0.6$ to represent the true
metadata balance of easy and hard digits. For this experiment, the random
method converges to a balance of 0.6, and random action converges to a
proportion of 0.5. The optimal proportion balance selected by GPAML is much
more consistent in this case than with the Spambase data. In particular, there
appears to be a narrow range of optimal proportion balances, ranging from
about 0.55 to 0.65. What is interesting here is that GPAML quickly identifies
that optimal proportion balance, and converges to it more quickly than either
competitor.

\subsection{RarePlanes}\label{sec:rareplanes}

The final dataset we consider is our motivating example, the RarePlanes data
\citep{shermeyer2021rareplanes}. This is a collection of 8,525 aerial images
of planes. The RarePlanes dataset contains many forms of metadata, but we
limit our focus to metadata describing the weather; we show that this metadata 
is suitable for our experiment in Appendix \ref{sup:rareplanesmeta}. In the original RarePlanes
dataset, this is divided into three categories: Clear Skies, Cloud Cover or
Haze, and Snow. For simplicity, we combine the first two categories into a
single category, Not Snow. YOLOV8 \citep{redmon2016you,yolov8} is used for the ML model.
YOLOV8 requires a user-specified validation set to use during model training. 
We sample a validation set of size $0.1N$, independent of $D_{N_i}$. For 
the purposes of training the GP, we treat any image in the validation set 
as part of the training set; i.e., the values of $N_{A_i}$ and  $N_{B_i}$ 
include images in the validation set.

It is not immediately obvious why we would want to utilize GPAML for 
this dataset. Although here we have immediate access to all data 
points, in the real world we would need to manually collect data at 
each step of the active learning process. In particular, for each new 
data acquisition, we would need to collect new aerial images. This 
requires launching an airplane to capture the images. This is very 
expensive, since each launch requires people, gas, and time. This 
could even require a flight into hostile territory, meaning we want 
to collect data as efficiently as possible.


This experiment is very computationally intensive.  
YOLOV8 requires GPUs, and even when using those GPUs each model fit is slow. 
On average, each ML model training took around 30 minutes for the smallest 
sample size, and 4 hours for the largest sample size; in the previous two experiments 
model training took fewer than 5 seconds. Without parallelization, one 
rep of the experiment would take over two months to run. Note that this two 
month figure assumes resources are available for the duration of the experiment. 
Since we ran this experiment on a shared supercomputing cluster, we often 
had to wait for resources to become available, increasing the time needed to 
run the experiment. Therefore, even after parallelization each rep of the 
experiment takes around a month to run. For this reason, we 
modified the experimental setup.

 Instead of fitting 100 reps, we fit three reps with
 three specific starting metadata balances: 25\% snow, 50\% snow, 
and 75\% snow. These three metadata 
balances are chosen to represent cases starting with balanced data, those 
starting with majority snow images, and those starting with majority non-snow 
images, respectively. We also consider different comparator methods:
\begin{itemize}
\item \textbf{GPAML}: Our method as described in Section \ref{sec:metalearn}.
\item \textbf{10\% snow}: New images are selected at true proportion balance.
\item \textbf{50\% snow}: New images are selected at 50/50 rate.
\item \textbf{All Snow}: All new images are snow images.
\item \textbf{All Not Snow}: All new images are not snow images.
\end{itemize}
These methods are chosen in an attempt to control the variability of the 
experiment. We hope to capture the ``best'' and ``worst'' acquisition decisions 
by looking at the Snow and Not Snow competitors, since we assume one metadata 
category is more valuable to the ML model than the other. However, we also 
include comparators which choose some images from each metadata category: 
50\% snow, representing balanced metadata adds, and 10\% snow, representing metadata 
added at the true proportion balance. The 10\% snow comparator is chosen 
because we saw in the Spambase and MNIST experiments that the optimal proportion  
balance is often equal to the true proportion balance.

\begin{figure}[ht!]
\includegraphics[scale=0.43, trim = {0, 15, 30, 20}, clip]{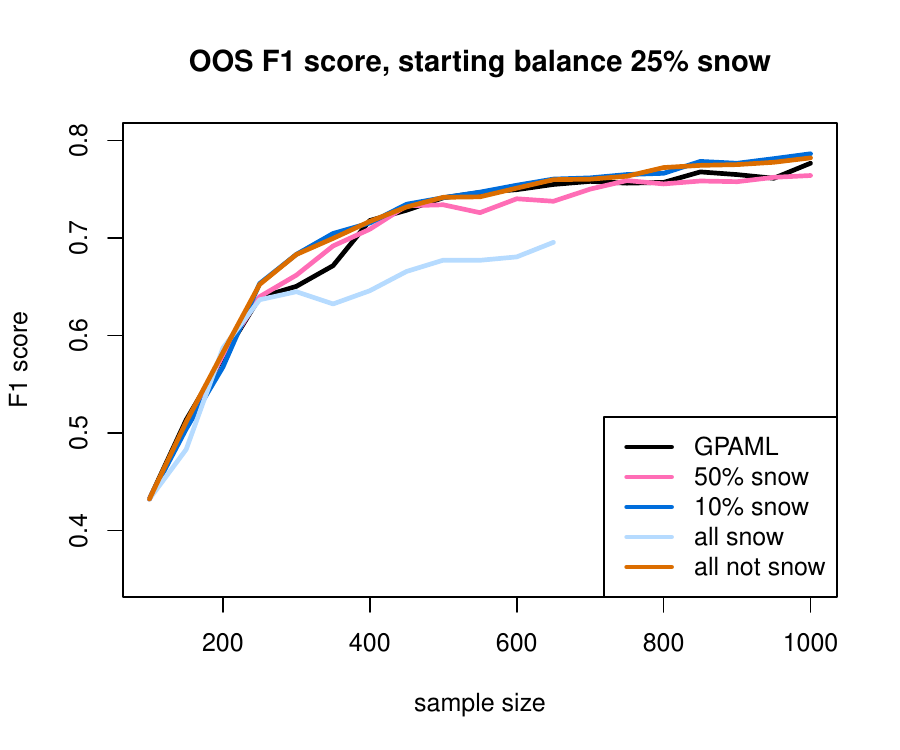}
\hfill
\includegraphics[scale=0.43, trim = {30, 15, 30, 20}, clip]{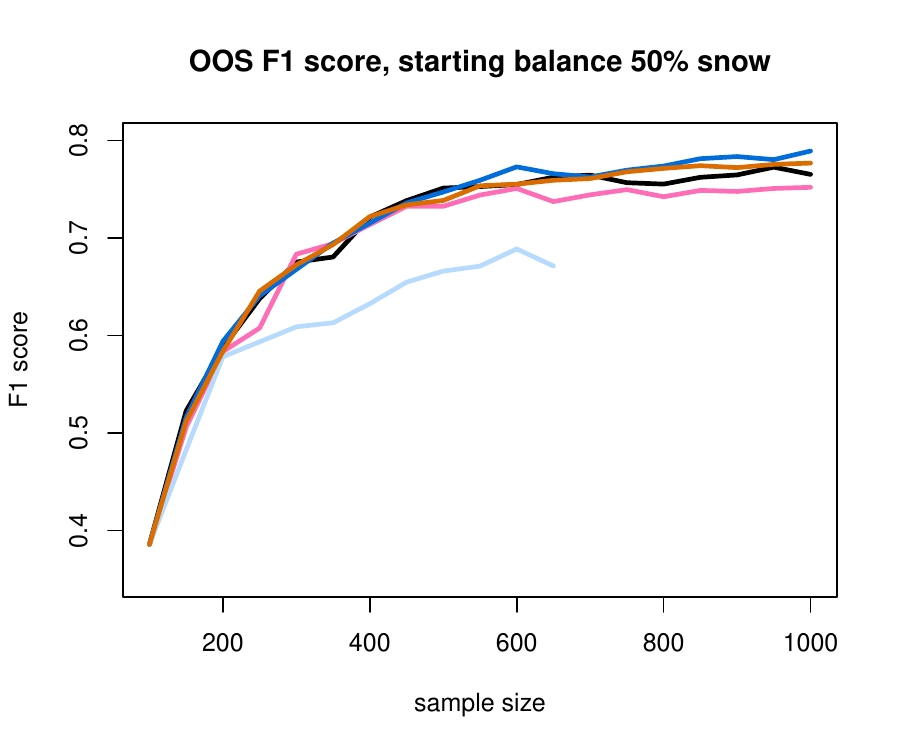}
\hfill
\includegraphics[scale=0.43, trim = {30, 15, 30, 20}, clip]{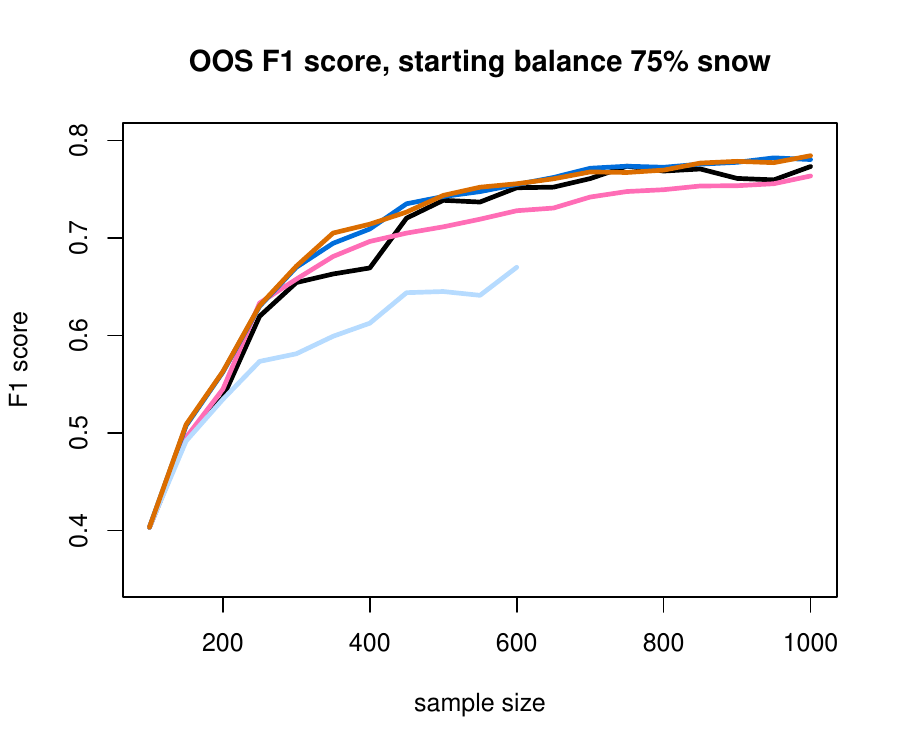}
\caption{ Out-of-sample performance for the 25\% snow {\em (left)},  
50\% snow {\em (center)}, and 75\% snow {\em (right)}  experiments.
}
\label{fig:rareplanesoos}
\end{figure}

For the RarePlanes experiment we began with $N_\mathrm{start}=100$ images, and
added $n=50$ new images at a time until reaching size $N_\mathrm{stop}=1000$.
For this experiment, we use F1 score to measure model accuracy. Out-of-sample of
performance of the three reps is shown in Figure \ref{fig:rareplanesoos}.
The three panels show, from left to right, the results of the 25\% snow, 
50\% snow, and 75\% snow starting metadata balances, respectively. We will comment 
on all three figures simultaneously since their results are similar. 

For each of the three starting metadata balances, notice that out-of-sample
model performance is roughly equivalent for all five competitors until size
$N=250$. Once reaching a size of $N=250$, the Snow competitor begins to show
much worse OOS performance than the other four competitors. This indicates
that having a large proportion of Snow images is suboptimal, with this
suboptimality worsening as the experiment continues. Moreover, note that the
Snow method stops prior to our final size of $N_\mathrm{stop}=1000$ because we
only have 600 Snow images available for $D_N$. The other four methods perform
comparably to each other for most of the experiment, except for a bit of
noise. However, only GPAML does not require pre-specified sampling proportions 
of the two metadata categories. We note that GPAML does not always make the 
optimal decision; note in particular the move from $N=300$ to $N=350$ in the 
75\% Snow experiment. We explore this suboptimality in Appendix 
\ref{sup:sampcompare}.

\begin{figure}[ht!]
\includegraphics[scale=0.41, trim = {0, 15, 15, 20}, clip]{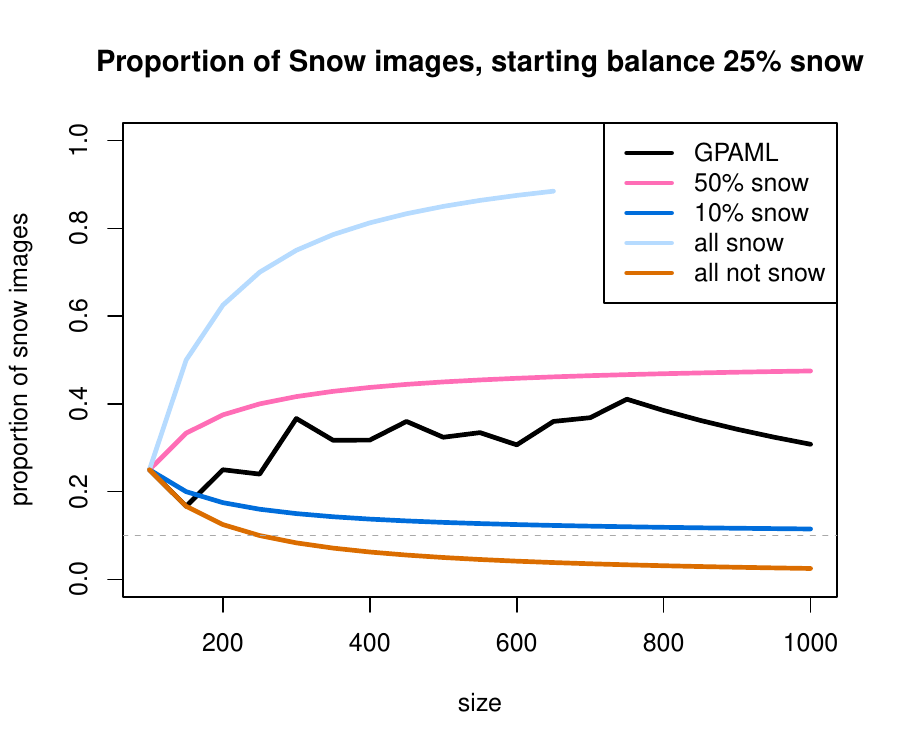}
\hfill
\includegraphics[scale=0.41, trim = {30, 15, 15, 20}, clip]{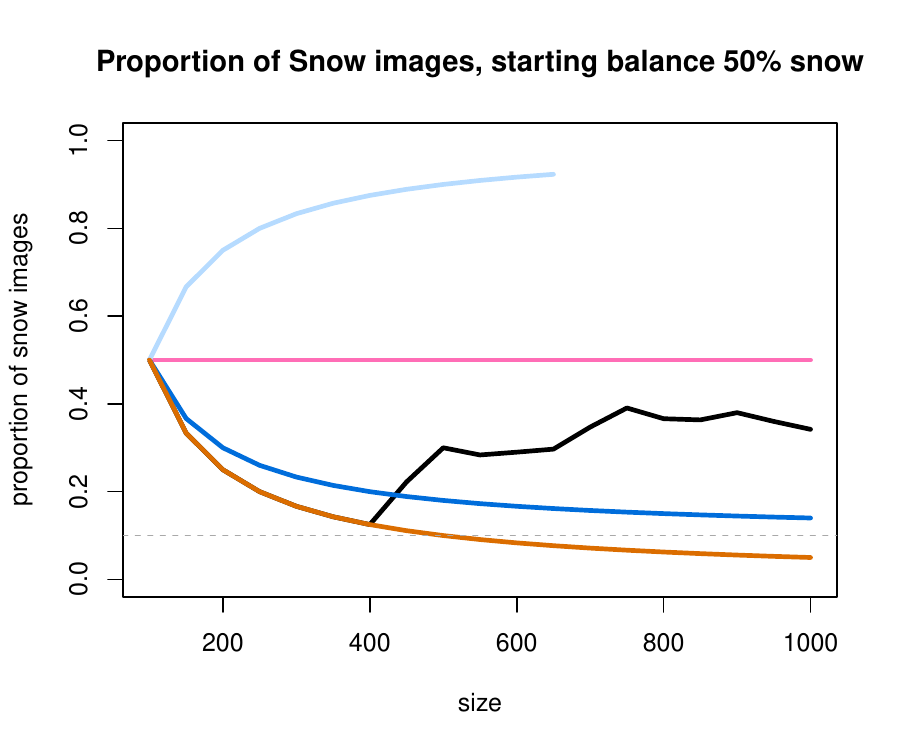}
\hfill
\includegraphics[scale=0.41, trim = {30, 15, 15, 20}, clip]{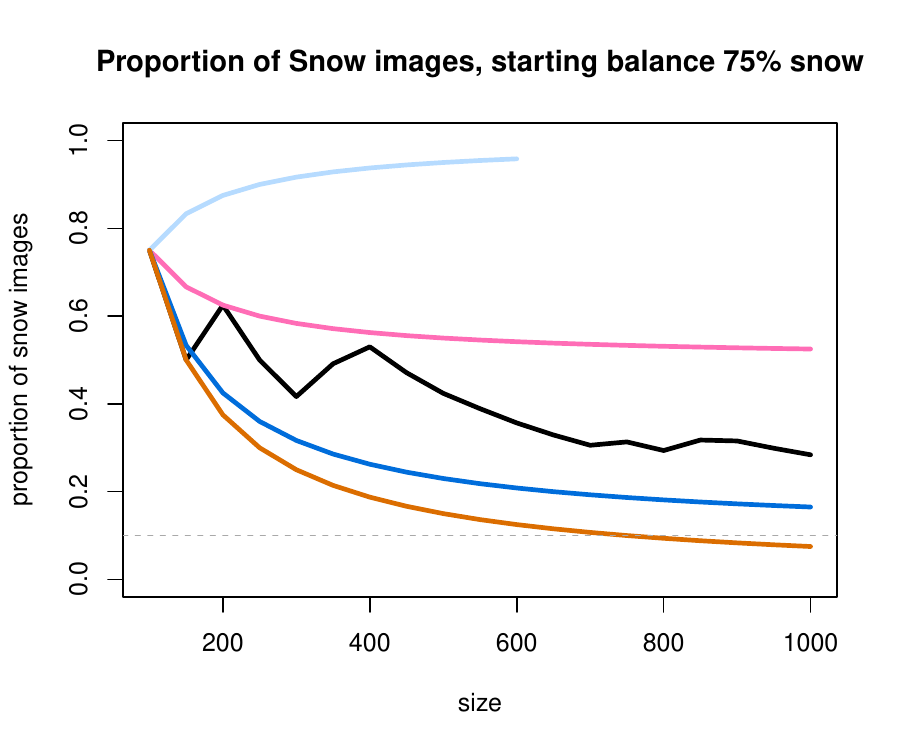}
\caption{Proportion balances for the 25\% snow {\em (left)},  
50\% snow {\em (center)}, and 75\% snow {\em (right)}  experiments.
}
\label{fig:rareplanesprop}
\end{figure}

For additional perspective, Figure \ref{fig:rareplanesprop} shows the
proportion balance for each of the three reps, relative to the Snow
balance.
They are shown in the same order as the previous plot. In each panel, a dashed
line is drawn horizontally at $y=0.1$ to indicate the true proportion balance
of Snow images. In these plots, the four non-GPAML competitors show smooth
changes in proportion balance, since we add the same balance of images at each
acquisition. So we focus on the results of GPAML. For each GPAML run, notice
that GPAML prefers, but does not exclusively choose, data points from the Not
Snow category. It is also able to self-correct after making a bad choice; in
the 75\% snow rep, it bounces between choosing lots of Snow and lots of Not
Snow images, eventually settling on a path that selects primarily Not Snow
images. These results match what we saw in the out-of-sample F1 results; that
competitors with primarily Not Snow images performed better (All Not Snow,
10\% snow). What is interesting is that the optimal proportion balance does
not seem to be converging to the true 10\% Snow balance; in all three cases,
the ending Snow percentage is around 30\%.

\section{Discussion}\label{sec:discussion}

We introduced a new idea for optimally selecting new data for ML models by
considering the metadata associated with the data points to be added. We call
this method GPAML, since we use a GP to learn the optimal metadata balance of
new data points. We found that GPAML performs at least as well as (and in some
cases better than) other data acquisition methods. In this scenario performing
as well as other data acquisition methods is suitable because other methods
may also find the optimal metadata balance. Most importantly, GPAML protects
against making data acquisitions that would hurt performance.

GPAML is restricted by the need to take subsamples from the existing data 
$D_N$. Because we asses the impact of metadata on model performance by 
taking subsamples from existing metadata categories, GPAML cannot estimate 
the impact of adding a data point from a new metadata category to the dataset. 
Additionally, if the number of data points in an existing metadata category is 
too small, it can be difficult for GPAML to estimate the impact of that category. 
Thus, for now we must begin with a dataset containing all metadata categories 
of interest, where the proportion balance is not extremely unbalanced. 

One possible area for future work is to extend GPAML to metadata with more
than two categories. The idea behind this is simple, although computational
expense quickly increases. To train a GP for GPAML, we were considering
$b=100$ unique metadata balances in two dimensions. To get a suitable accuracy
surface in higher dimension, we would likely need to take even more samples.
Since each added rep requires fitting another ML model, the computational
expense can quickly become unmanageable.

We are also interested in developing strategies that would mitigate the 
large computational cost of GPAML. We believe there may be a way 
to do this by selecting the training points $X^{(N)}$ for the GP in a more 
intelligent way. In particular, we want to develop a version of GPAML 
which reuses subsamples from previous iterations, so we do not need 
to fit 100 ML models at each iteration. One simple solution is to only 
collect new subsamples in the unexplored portions of the space. 
However, this would mean new images are never incorporated into 
the original space. We believe this idea could be modified by supplementing 
the new subsamples with a few additional subsamples in the original 
space, but exploring that idea represents future work.

A current shortcoming of this method is the assumption that we know the
underlying proportion balance of the metadata, and using that as the
proportion balance for the test set $D_{N_i}^\mathrm{test}$. We made this
decision after learning that random selection of the proportion selection of
$D_{N_i}^\mathrm{test}$ yielded suboptimal results. In particular, by choosing
$D_{N_i}^\mathrm{test}$ at random, its proportion balance (over many reps)
converges to the proportion balance of $D_N$. We found that this can lead to a
vicious cycle, where the current proportion balance is chosen as optimal for
each acquisition; i.e., GPAML is not learning the true optimal metadata
balance. We needed a way to avoid this vicious cycle. Assuming the underlying
proportion balance is simple, but not always realistic. We would like to find
a different way to specify the proportion balance of $D_{N_i}^\mathrm{test}$
that does not rely on this assumption.

\subsection*{Acknowledgements}
This material is based upon work supported, in whole or in part, by the 
U.S.~Department of Defense, Director, Operational Test and Evaluation
 (DOT\&E) through the Office of the Assistant Secretary of Defense for 
 Research and Engineering (ASD(R\&E)) under contracts HQ003419D0003 
 and HQ003424D0023. The Systems Engineering Research Center (SERC) 
 is a federally funded University Affiliated Research Center managed by 
 Stevens Institute of Technology. Any views, opinions, findings and conclusions 
 or recommendations expressed in this material are those of the author(s) and 
 do not necessarily reflect the views of the United States Department of Defense 
 nor DOT\&E nor ASD(R\&E).

\subsection*{Declaration of Interest}
The authors report no conflict of interest.

\singlespacing
\bibliographystyle{apalike}
\bibliography{references}

\newpage
\begin{center}
{\large\bf SUPPLEMENTARY MATERIAL}
\end{center}
\appendix

\section{Metadata Preprocessing}\label{sup:meta}

In order for a metadata category to be suitable for GPAML, or any other metadata-based 
acquisition strategy, that category needs to have a meaningful impact on model 
performance. Here we demonstrate that our chosen metadata variables are expected 
to have an impact on performance.

\subsection{Spambase}\label{sup:spammeta}

\begin{figure}[ht!]
\includegraphics[scale=0.55, trim = {0, 45, 30, 20}, clip]{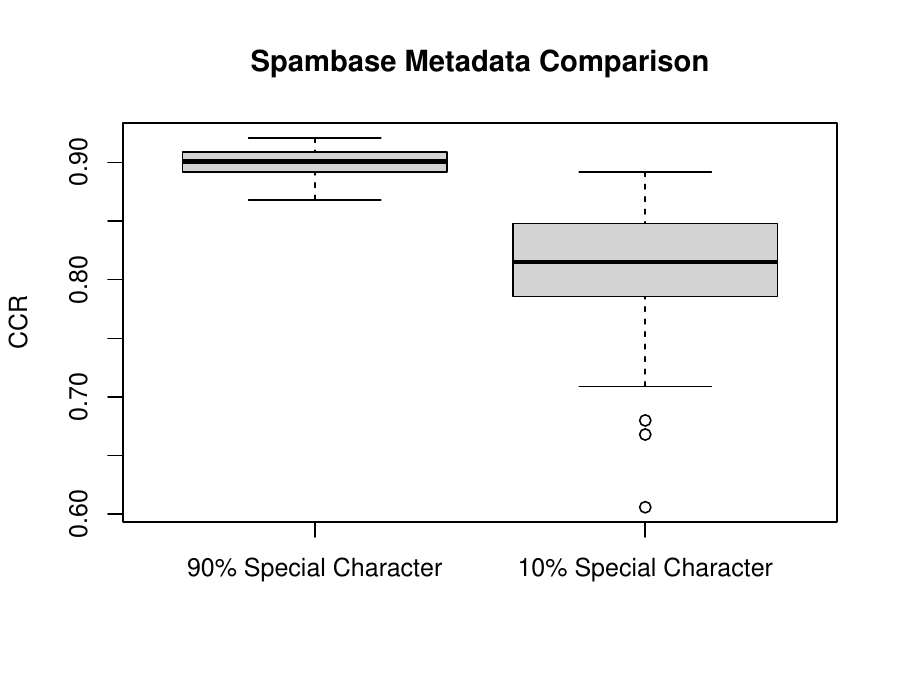}
\hspace{1cm}
\includegraphics[scale=0.55, trim = {0, 45, 30, 20}, clip]{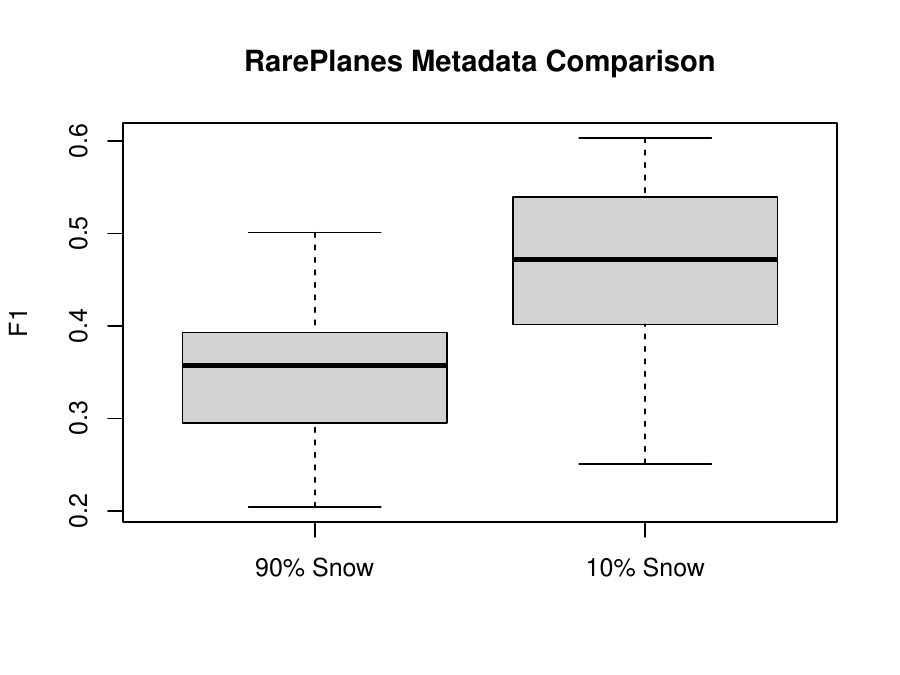}
\caption{{\em Left:} OOS CCR for Spambase metadata comparison; {\em right:} 
OOS F1 score for RarePlanes metadata comparison.}
\label{fig:meta1}
\end{figure}

For the Spambase dataset we used metadata separating emails into those with no special characters, 
and those with at least one special character. To check whether this metadata affects model performance 
we performed a simple experiment comparing datasets comprised primarily of emails with no special 
characters to datasets comprised primarily of emails with at least one special character. In particular, 
for each metadata category, we randomly selected 90 emails from that metadata category, and 10 emails 
from the other metadata category. We then evaluated this dataset on a holdout set of 500 emails. This 
was repeated 100 times for each metadata category, creating the boxplot in the left panel of Figure \ref{fig:meta1}.
Here we see that, in general, it is beneficial to collect emails with at least one special character. Since 
there is a difference in model performance for each metadata category, it is suitable for GPAML.

\subsection{MNIST}\label{sup:mnistmeta}

\begin{figure}[ht!]
\begin{subfigure}[c]{.6\textwidth}
\centering
\footnotesize
\begin{tabular}{ c|cccccccccc|c} 
 & 0 & 1 & 2 & 3 & 4 & 5 & 6& 7 &8 & 9  & CCR\\ 
\hline
0&92&0&2&1&0&2&2&0&1&3 & 0.89\\ 
1&0&113&1&0&0&0&0&0&0&0&0.99\\
2&0&2&65&0&5&0&2&4&1&9&0.74\\
3&0&1&5&74&0&4&1&4&9&2&0.74\\
4&0&1&0&0&89&1&0&0&9&3&0.86\\
5&0&1&0&19&4&47&5&2&10&6&0.5\\
6&1&2&1&3&8&0&71&0&0&1&0.82\\
7&0&1&3&0&0&2&0&96&3&17&0.79\\
8&0&2&2&2&0&5&0&1&93&3&0.86\\
9&0&0&0&2&11&4&0&10&11&43&0.53\\
\hline
CCR& 0.99 & 0.92 & 0.82 & 0.73 & 0.76 & 0.72 & 0.88 & 0.82 & 0.68 & 0.49
  \end{tabular}
\end{subfigure}
\hfill
\begin{subfigure}[c]{.3\textwidth}
\centering
\includegraphics[scale=0.4, trim = {0, 45, 30, 20}, clip]{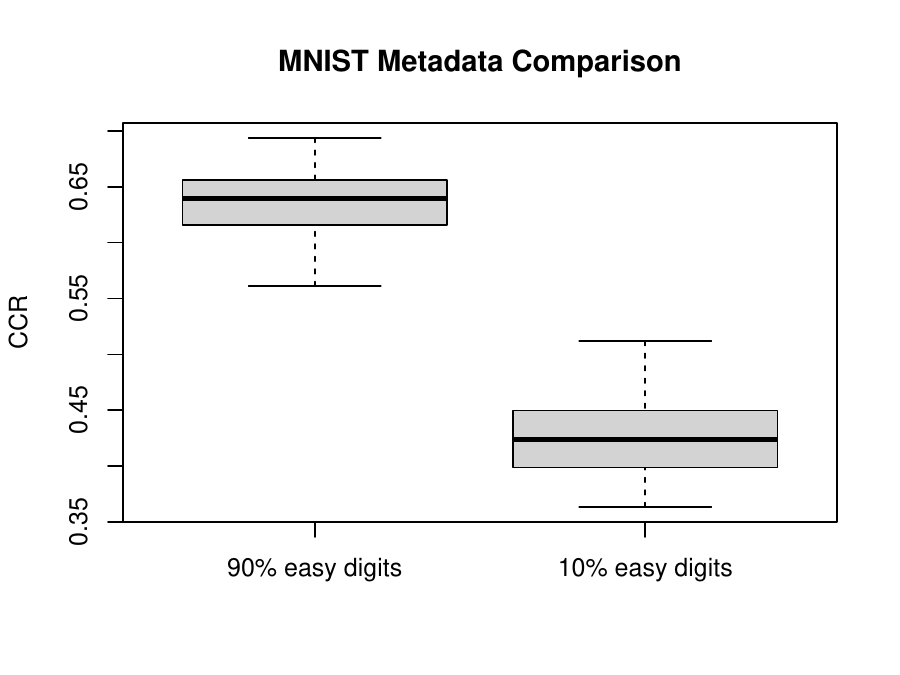}
\end{subfigure}
\caption{{\em Left:} Confusion matrix for a single rep of the MNIST experiment; {\em right:} 
OOS CCR for MNIST metadata comparison.}
\label{fig:meta2}
\end{figure}

For the MNIST dataset, we used metadata which separated digits into those easy to categorize 
and those difficult to categorize. By looking at confusion matrices comparing the true and predicted 
digits for a single run of the NN, we can learn which MNIST digits are easy and difficult to categorize. 
We looked at the confusion matrices produced by several runs of the model to learn: 
(i) digits that are easy and difficult to categorize and (ii) digits that are frequently confused with 
each other.  An example of a confusion matrix is shown in \ref{fig:meta2}. 
From this plot, 
we can learn which digits are frequently misclassified, and which are frequently predicted incorrectly. 
We attempt to summarize these two pieces of information (i.e., the final row and column of the 
confusion matrix), grouping together digits that are frequently confused with each other.
 Summarizing all this information, we placed digits 
0, 1, 4, 5, 6, and 7 and in the easy category, and digits 2, 3, 8, and 9 in the hard category.

To double check that we have created metadata suitable for GPAML, we performed an experiment 
similar to that in Section \ref{sup:spammeta}. In particular, for each metadata category we randomly 
selected 90 images from that metadata category and 10 images from the opposite category. The 
model performance was evaluated on a holdout set of 1000 images. We repeated this process 100 
times for each metadata category to create the boxplot in the right panel of Figure \ref{fig:meta2}.

\subsection{RarePlanes}\label{sup:rareplanesmeta}

For the RarePlanes dataset we used metadata describing the weather in which an image was taken: 
snowy, or not snowy. To see whether this metadata affects model performance we performed an 
experiment similar to the one in Section \ref{sup:spammeta}. In particular, for each metadata 
category, we randomly selected 90 images from that metadata category and 10 images from the 
opposite category. The model performance was evaluated on a holdout set of 1000 images. This 
process was repeated 100 times for each metadata category, creating the boxplot in the right panel of Figure 
\ref{fig:meta1}. Here we see that, in general, it is beneficial to collect more images 
in the Not Snow category. This difference in model performance makes it suitable for GPAML.

\section{Effect of different samples}\label{sup:sampcompare} 

The choice of optimal metadata balance by GPAML depends on the subsamples selected 
during the metadata balance varying experiment. To explore this further, 
we look at the impact of different subsamples at one acquisition of the 
RarePlanes experiment. In particular, we explore the experiment starting with 75\% Snow 
and look at the move from $N=300$ to $N=350$. We chose this move because GPAML 
suddenly underperforms its competitors (see Figure \ref{fig:rareplanesoos}). This is caused 
by selecting primarily Snow images (see Figure \ref{fig:rareplanesprop}), which is a 
suboptimal decision. We complete a simple experiment to explore the effect of subsamples 
on the acquisition decision. Instead of 
our typical $b=100$ unique metadata balances chosen, we simulated a total of
 $b=250$ unique metadata balances; this includes the 100 subsamples selected 
 in the original experiment. 

\begin{figure}[ht!]
\includegraphics[scale=0.42, trim = {0, 15, 30, 20}, clip]{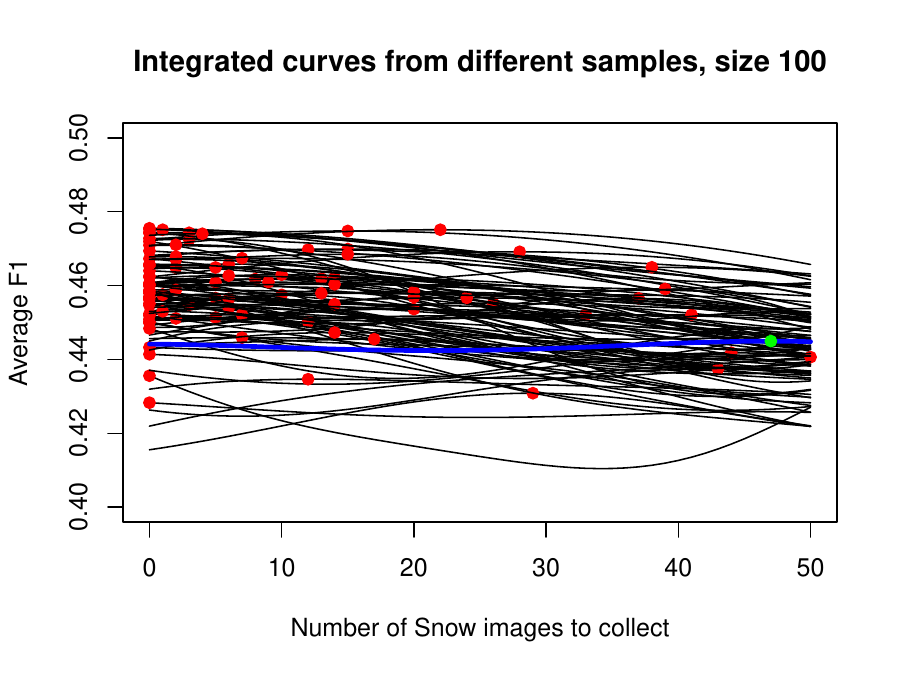}
\hfill
\includegraphics[scale=0.42, trim = {15, 15, 30, 20}, clip]{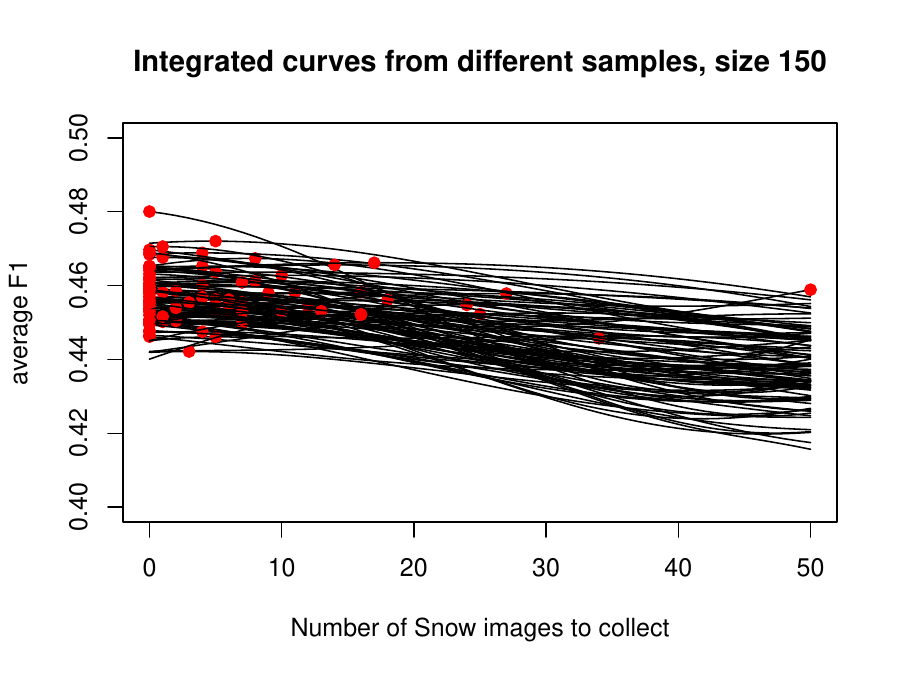}
\hfill
\includegraphics[scale=0.42, trim = {15, 15, 30, 20}, clip]{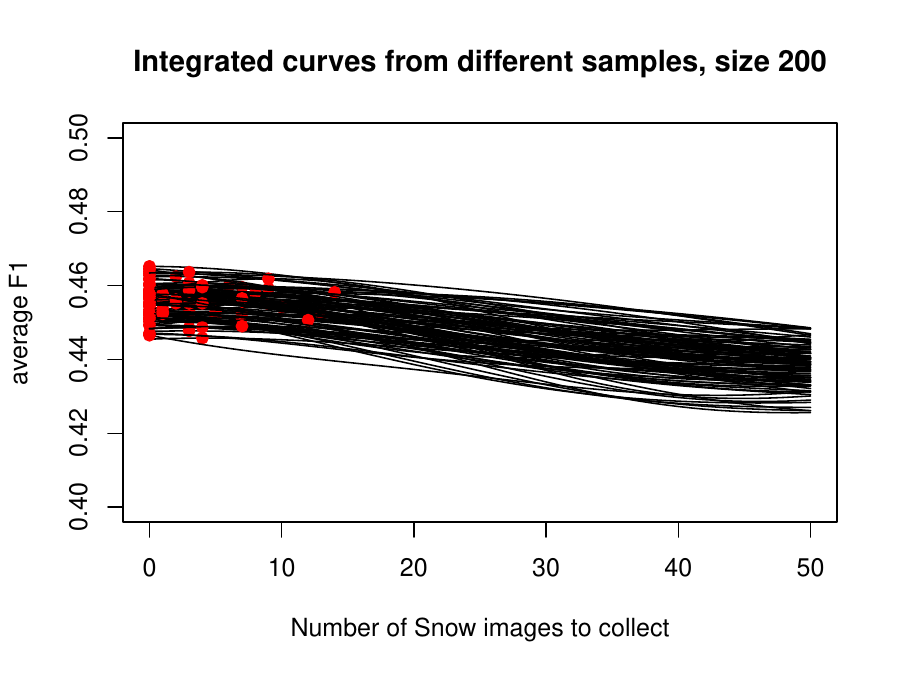}
\caption{Integrated GPAML curve from 100 random samples of points. Acquisition choices are 
shown as red dots. The blue line shows our original acquisition line, with a green point denoting 
the original acquisition decision.}
\label{fig:sampvary}
\end{figure}

From this collection of unique metadata balances, we designed three experiments to explore 
the effect of sample collection on the decision made by GPAML. In particular, from these 
$b=250$ unique metadata balances, we took random samples of sizes 100, 150, and 200, 
and used the selected samples to fit the GP. We took 100 random samples for each sample 
size. The acquisition decision line for each GP 
is shown in Figure \ref{fig:sampvary} for samples of sizes 100, 150, and 200, from left 
to right.
In this figure, we plot the integrated model performance learned from each GP as a black 
line, with a red dot denoting the acquisition decision. We also include the original sample 
of size 100 (i.e., the one shown in Figures \ref{fig:rareplanesoos} and \ref{fig:rareplanesprop}) 
as a blue line, with a green dot denoting the acquisition decision.

From the results in Section \ref{sec:rareplanes}, we learned that in general, any acquisition 
decision that chooses a majority of Not Snow images is a good decision. Using this criterion, 
we see that GPAML makes a good decision around 90\% of the time with samples of size 
100, around 98\% of the time for size 150, and 100\% of the time for size 200. It is 
unsurprising that collecting more samples 
provides more consistent results. However, given the extra computational cost of collecting 
50 more samples, we are satisfied with a method that makes the correct decision 90\% of 
the time. We also note that the sub-optimal decision which originally interested us was a 
particularly unlucky decision; only four other samples of size 100 made a decision as bad 
or worse. We also note that, although GPAML made a bad decision at size $N=300$, 
it was able to correct itself in the next step, at $N=350$.

\end{document}